\documentclass[10pt,twocolumn,letterpaper]{article}

\makeatletter
\@namedef{ver@everyshi.sty}{}
\makeatother
\usepackage[pagenumbers]{cvpr} 

\usepackage{graphicx}
\usepackage{amsmath}
\usepackage{amssymb}
\usepackage{booktabs}
\usepackage{esint}
\usepackage{svg}
\usepackage{zref-base}
\usepackage{atveryend}

\usepackage[pagebackref,breaklinks,colorlinks]{hyperref}

\usepackage[capitalize]{cleveref}
\crefname{section}{Sec.}{Secs.}
\Crefname{section}{Section}{Sections}
\Crefname{table}{Table}{Tables}
\crefname{table}{Tab.}{Tabs.}

\def\cvprPaperID{7829} 
\def\confName{CVPR}
\def\confYear{2023}

\newcommand{\name}{Exact-NeRF}
\newcommand{\ihat}{\hat{\textbf{\i}}}

\makeatletter
\AfterLastShipout{%
  \if@filesw   
    \begingroup
      \zref@setcurrent{default}{\the\value{equation}}%
      \zref@wrapper@immediate{%
        \zref@labelbyprops{eq-last}{default}%
      }%
    \endgroup
  \fi
}
\makeatother

\begin{document}

\title{{\name}: An Exploration of a Precise Volumetric \\ Parameterization for Neural Radiance Fields}

\author{Brian K. S. Isaac-Medina\textsuperscript{1}, Chris G. Willcocks\textsuperscript{1}, Toby P. Breckon\textsuperscript{1, 2}\\
Department of \{\textsuperscript{1}Computer Science, \textsuperscript{2}Engineering\}, Durham University, UK\\
{\tt\small \{brian.k.isaac-medina, christopher.g.willcocks, toby.breckon\}@durham.ac.uk}
}

\maketitle

\begin{abstract}
  \noindent
  Neural Radiance Fields (NeRF) have attracted significant attention due to their ability to synthesize novel scene views with great accuracy. However, inherent to their underlying formulation, the sampling of points along a ray with zero width may result in ambiguous representations that lead to further rendering artifacts such as aliasing in the final scene. 
  To address this issue, the recent variant mip-NeRF proposes an Integrated Positional Encoding (IPE) based on a conical view frustum. Although this is expressed with an integral formulation, mip-NeRF instead approximates this integral as the expected value of a multivariate Gaussian distribution.
  This approximation is reliable for short frustums but degrades with highly elongated regions, which arises when dealing with distant scene objects under a larger depth of field. In this paper, we explore the use of an exact approach for calculating the IPE by using a pyramid-based integral formulation instead of an approximated conical-based one. We denote this formulation as {\name} and contribute the first approach to offer a precise analytical solution to the IPE within the NeRF domain. Our exploratory work illustrates that such an exact formulation ({\name}) matches the accuracy of mip-NeRF and furthermore provides a natural extension to more challenging scenarios without further modification, such as in the case of unbounded scenes. Our contribution aims to both address the hitherto unexplored issues of frustum approximation in earlier NeRF work and additionally provide insight into the potential future consideration of analytical solutions in future NeRF extensions.
\end{abstract}


\begin{figure}[t]
\centering
\subfloat{\includegraphics[width=\linewidth]{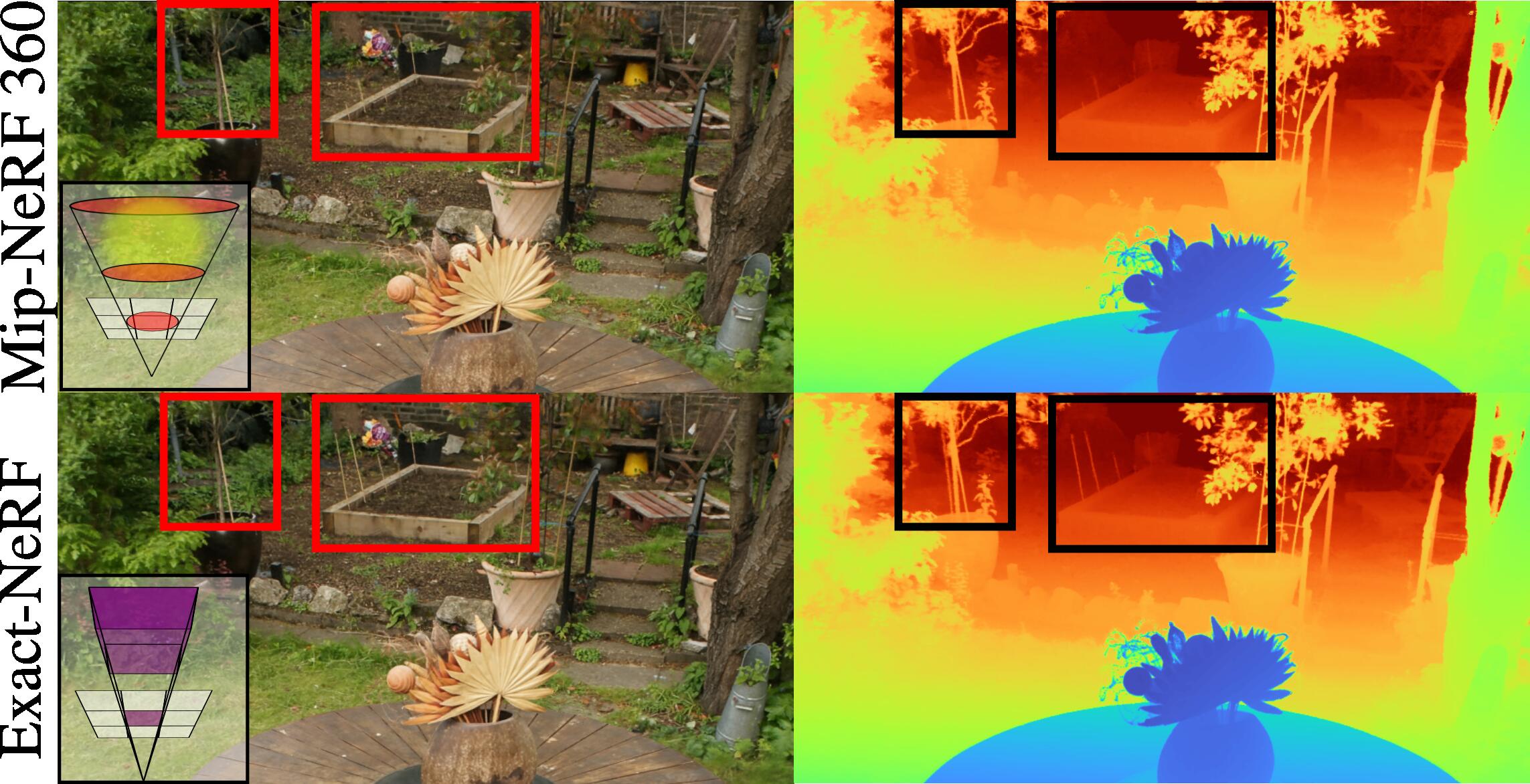}}
\caption{Comparison of {\name} (ours) with mip-NeRF 360 \cite{barron2022mip}. Our method is able to both match the performance and obtain superior depth estimation over a larger depth of field.}
\label{fig:front-image}
\end{figure}

\section{Introduction}
\noindent
Novel view synthesis is a classical and long-standing task in computer vision that has been thoroughly re-investigated via recent work on Neural Radiance Fields (NeRF) \cite{mildenhall2021nerf}. NeRF learns an implicit representation of a 3D scene from a set of 2D images via a Multi-Layer Perceptron (MLP) that predicts the visual properties of 3D points uniformly sampled along the viewing ray given its coordinates and viewing direction. This parameterization gives NeRF the dual ability to both represent 3D scenes and synthesize unseen views. In its original formulation, NeRF illustrates strong reconstruction performance for synthetic datasets comprising object-centric scenes and no background (bounded) and forward-facing real-world scenes. Among its applications, NeRF has been used for urban scene representation \cite{tancik2022block,xiangli2021citynerf,rematas2022urban}, human body reconstruction \cite{SMPL:2015,coronafigueroa2022mednerf}, image processing \cite{mildenhall2022rawnerf,huang2022hdr,ma2022deblur} and physics \cite{guan2022neurofluid,levis2022gravitationally}.

Nonetheless, the underlying sparse representation of 3D points learnt by the MLP may cause ambiguities that can lead to aliasing and blurring. To overcome these issues, Barron \etal proposed mip-NeRF \cite{barron2021mip}, an architecture that uses cone tracing instead of rays. This architecture encodes conical frustums as the inputs of the MLP by approximating the integral of a sine/cosine function over a region in the space with a multivariate Gaussian. This re-parameterization notably increases the reconstruction quality of multi-scale datasets. However, this approximation is only really valid for bounded scenes, where the conic frustums do not suffer from large elongations attributable to a large depth of field within the scene. 


The NeRF concept has been extended to represent increasingly difficult scenes. For instance, mip-NeRF 360 \cite{barron2022mip} learns a representation of unbounded scenes with a central object by giving more capacity to points that are near the camera, modifying the network architecture and introducing a regularizer that penalizes `floaters' (unconnected depth regions in free space) and other small unconnected regions. In order to model distant regions, mip-NeRF 360 transforms the multivariate Gaussians with a contraction function. This modification allows a better representation and outperforms standard mip-NeRF for an unbounded scenes dataset. However, the modification of the Gaussians requires attentive analysis to encode the correct information in the contracted space, 
which includes the linearization of the contraction function to accommodate the Gaussian approximations. This leads to a degraded performance of mip-NeRF 360 when the camera is far from the object. Additionally, mip-NeRF 360 struggles to render thin structures such as tree branches or bicycle rays.

Motivated by this, we present {\name} as an exploration of an alternative exact parameterization of underlying volumetric regions that are used in the context of mip-NeRF (\cref{fig:front-image}). We propose a closed-form volumetric positional encoding formulation (\cref{sec:method}) based on pyramidal frustums instead of the multivariate Gaussian approximation used by mip-NeRF and mip-NeRF 360. {\name} matches the performance of mip-NeRF on a synthetic dataset, but gets a sharper reconstruction around edges. Our approach can be applied without further modification to the contracted space of mip-NeRF 360. Our naive implementation of {\name} for the unbounded scenes of mip-NeRF 360 has a small decrease in performance, but it is able to get cleaner reconstructions of the background. Additionally, the depth map estimations obtained by {\name} are less noisy than mip-NeRF 360. Our key contribution is the formulation of a general integrated positional encoding framework that can be applied to any shape that can be broken into triangles (\ie, a polyhedron). We intend that our work serves as a motivation to investigate different shapes and analytical solutions of volumetric positional encoding. The code is available at \url{https://github.com/KostadinovShalon/exact-nerf}.

\section{Related Work}
\noindent
Already numerous work has focused on improving NeRF since its original inception \cite{mildenhall2021nerf}, such as decreasing the training time \cite{mipnerfrgbd, Hu_2022_CVPR, Deng_2022_CVPR, Fridovich-Keil_2022_CVPR, recursive-nerf}, increasing the synthesis speed \cite{Hedman_2021_ICCV, Yu_2021_ICCV, sitzmann2021light}, reducing the number of input images \cite{Niemeyer2021Regnerf} and improving the rendering quality \cite{lin2021barf, barron2021mip, barron2022mip, zhang2020nerf++, martin2021nerf, verbin2022refnerf}. With the latter, one of the focuses is to change the positional encoding to account for the volumetric nature of the regions that contribute to pixel rendering \cite{barron2021mip, barron2022mip}. 
\\ 
\subsection{Positional Encoding}
\label{sec:pe}
\noindent
NeRF uses a positional encoding (PE) on the raw coordinates of the input points in order to induce the network to learn higher-frequency features \cite{spectralbias}. However, the sampled points in NeRF are intended to represent a region in the volumetric space. This can lead to ambiguities that may cause aliasing. In this sense, mip-NeRF \cite{barron2021mip} uses a volumetric rendering by casting cones instead of rays, changing the input of the MLP from points to cone frustums. These regions are encoded using an integrated positional encoding (IPE), which aims to integrate the PE over the cone frustums. Given that the associated integral has no closed-form solution, they formulate the IPE as the expected value of the positional encoding in a 3D Gaussian distribution centred in the frustum. The IPE reduces aliasing by reducing the ambiguity of single-point encoding. Mip-NeRF 360 \cite{barron2022mip} uses a contracted space representation to extend the mip-NeRF parameterization to 360° unbounded scenes, since they found that the approximation given in mip-NeRF degrades for elongated frustums which arise in the background. Additionally, and similar to DONeRF \cite{neff2021donerf}, mip-NeRF 360 samples the intervals of the volumetric regions using the inverse of the distance in order to assign a bigger capacity to nearer objects. By contrast, in this work we explore the use of pyramid-based frustums in order to enable an exact integral formulation of the IPE which can be applied for both bounded and unbounded scenes alike. 

\subsection{NeRF and  Mip-NeRF parameterization}
\label{sec:nerf}
\noindent
NeRF uses an MLP $f$ with parameters $\Theta$ to get the colour $\mathbf{c} \in \mathbb{R}^3$ and density $\sigma \in \left[0, +\infty \right)$ given a point $\mathbf{x} \in \mathbb{R}^3$ and a viewing direction $\hat{\mathbf{v}} \in S^2$, where $S^2$ is the unit sphere, such that:
\begin{equation}
    \left( \mathbf{c}, \sigma \right) = f(\mathbf{x}, \hat{\mathbf{v}}; \Theta )\,,
\end{equation}
\noindent
whereby NeRF uses the positional encoding
\begin{align}
\label{eq:pe}
\begin{split}
    \gamma(\mathbf{x}) = &\left[\sin(2^0\mathbf{x}), \hdots, \sin(2^{L-1}\mathbf{x}), \right. \\
    &\left. \cos(2^0\mathbf{x}), \hdots, \cos(2^{L-1}\mathbf{x}) \right]^\top,
\end{split}
\end{align}
\noindent
and $\gamma: \mathbb{R} \rightarrow \mathbb{R}^{2L}$ is applied to each coordinate of $\mathbf{x}$ and each component of $\hat{\mathbf{v}}$ independently. 
\begin{figure*}[t!]
\centering
\subfloat[]{\includegraphics[width=0.45\linewidth]{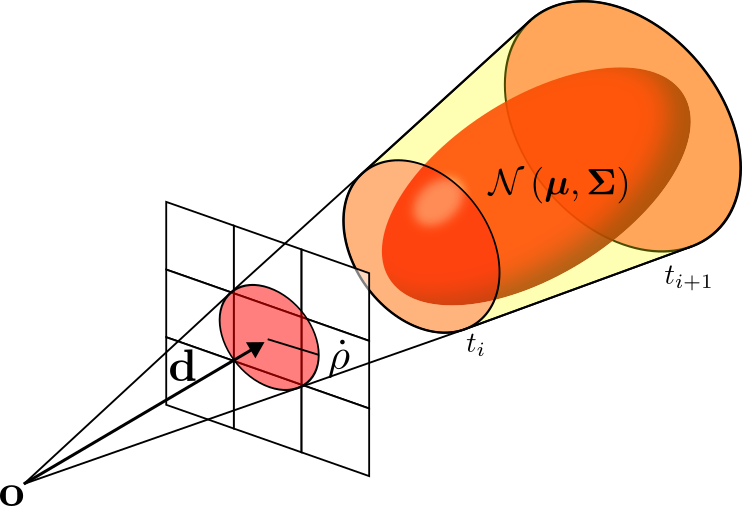} \label{cone-fig}}
\subfloat[]{\includegraphics[width=0.45\linewidth]{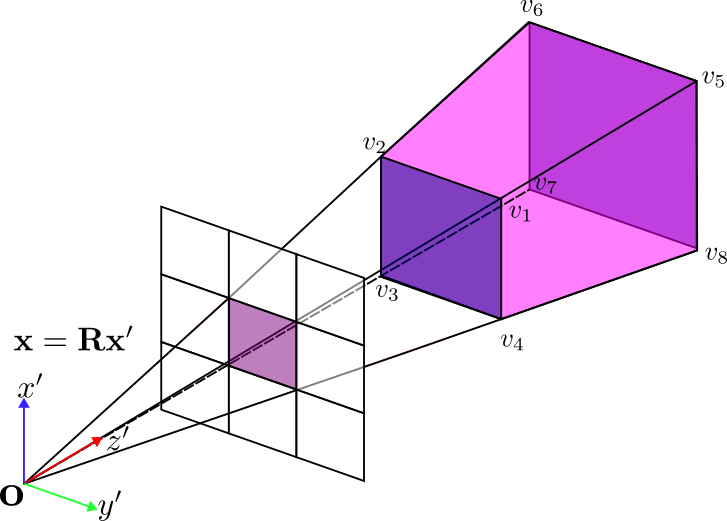} \label{pyramid-fig}}
\caption{Cone and pyramid tracing for volumetric NeRF parameterizations. (a) Mip-NeRF \cite{barron2021mip} uses cone frustums to parameterize a 3D region. Since the IPE of these frustums does not have a closed-form solution, it is approximated by modelling the frustum as a multivariate Gaussian. (b) {\name} casts a square pyramid instead of a cone, allowing for an exact parameterization of the IPE by using the vertices $v_i$ of the frustum and the pose parameters $\mathbf{o}$ and $\mathbf{R}$.}
\label{fig:frustums}
\end{figure*}
The sampling strategy of NeRF consists of sampling random points along the ray that passes through a pixel. This ray is represented by $\mathbf{r}(t) = t \mathbf{d} + \mathbf{o}$, where $\mathbf{o}$ is the camera position and $\mathbf{d}$ is the vector that goes from the camera centre to the pixel in the image plane. The ray is divided into $N$ intervals and the points $\mathbf{r}(t_i)$ are drawn from a  uniform distribution over each interval, such that:
\begin{equation}
\label{eq:sampling_t}
    t_i \sim \mathcal{U}\left[t_n + \frac{i-1}{N}(t_f - t_n), t_n + \frac{i}{N}(t_f - t_n)\right]\,,
\end{equation}
\noindent
where $t_n$ and $t_f$ are the near and far planes. In this sense, the colour and density of each point over the ray are obtained by $(\mathbf{c}_i, \sigma_i) = f(\gamma(\mathbf{r}(t_i)), \gamma(\mathbf{d}/\lVert \mathbf{d} \rVert); \Theta)$.

Finally, the pixel colour $\hat{C}(\mathbf{r})$ is obtained using numerical quadrature,
\begin{equation}
\begin{split}
    \hat{C}(\mathbf{r}) = \sum^N_{i=1} T_i (1 - \exp(-\sigma_i \delta_i))\\
    T_i= \exp \left( - \sum_{j=1}^{i-1} \sigma_j \delta_j \right)\,,
\end{split}
\end{equation}
where $\delta_i = t_{i+1}-t_i$. This process is carried out hierarchically by using coarse $\hat{C}_c$ and fine $\hat{C}_f$ samplings, where the 3D points in the latter are drawn from the PDF formed by the weights of the density values of the coarse sampling. The loss is then the combination of the mean-squared error of the coarse and fine renderings for all rays $\mathbf{r} \in \mathcal{R}$, \ie, 
\begin{equation}
    \mathcal{L} = \sum_{\mathbf{r} \in \mathcal{R}} \lVert \hat{C}_c(\mathbf{r}) - C(\mathbf{r}) \rVert_2^2 + \lVert \hat{C}_f(\mathbf{r}) - C(\mathbf{r}) \rVert_2^2\,.
\end{equation}
Here we find mip-NeRF \cite{barron2021mip} is similar to NeRF, but it utilises cone tracing instead of ray tracing. This change has the direct consequence of replacing ray intervals by conical frustums $F(\mathbf{d}, \mathbf{o},\dot{\rho}, t_i, t_{i+1})$, where $\dot{\rho}$ is the radius of the circular section of the cone at the image plane (\cref{fig:frustums}a). This leads to the need for a new positional encoding that summarizes the function in \cref{eq:pe} over the region defined by the frustum. The proposed IPE is thus given by:
\begin{equation}
    \label{eq:ipe}
    \gamma_I(\mathbf{d}, \mathbf{o},\dot{\rho}, t_i, t_{i+1}) = \frac{\iiint_F \gamma(\mathbf{x})dV}{\iiint_F dV}\,.
\end{equation}

Since the integral in the numerator of \cref{eq:ipe} has no closed-form solution, mip-NeRF proposes to approximate it by considering the cone frustums as multivariate Gaussians. Subsequently, the approximated IPE $\gamma^*$ is given by:
\begin{align}
\label{eq:approx-ipe}
\begin{split}
        \gamma^*&(\boldsymbol{\mu}, \boldsymbol{\Sigma}) = \mathbb{E}_{x \sim \mathcal{N}(\mathbf{P} \boldsymbol\mu, \mathbf{P} \mathbf{\Sigma} \mathbf{P}^\top)} \left[ \gamma(\mathbf{x}) \right] \\
        &= 
        \begin{bmatrix}
        \sin(\mathbf{P} \boldsymbol{\mu}) \circ \exp(-(1/2) \text{diag}(\mathbf{P} \mathbf{\Sigma} \mathbf{P}^\top)) \\
        \cos(\mathbf{P} \boldsymbol{\mu}) \circ \exp(-(1/2) \text{diag}(\mathbf{P} \mathbf{\Sigma} \mathbf{P}^\top))
        \end{bmatrix}\,,
\end{split}
\end{align}
where $\boldsymbol{\mu}=\mathbf{o} + \mu_t \mathbf{d}$ is the centre of the Gaussian for a frustum defined by $\mathbf{o}$ and $\mathbf{d}$ with mean distance along the ray $\mu_t$, $\mathbf{\Sigma}$ is the covariance matrix, $\circ$ denotes element-wise product and:
\begin{equation}
\setlength\arraycolsep{3pt}
\mathbf{P} = 
    \begin{bmatrix}
    1 & 0 & 0 & 2 & 0 & 0 &  & 2^{L-1} & 0 & 0 \\
    0 & 1 & 0 & 0 & 2 & 0 & \hdots & 0 & 2^{L-1} & 0 \\
    0 & 0 & 1 & 0 & 0 & 2 &  & 0 & 0 & 2^{L-1}
    \end{bmatrix}^\top\,.
\end{equation}
\noindent
This formulation was empirically shown to be accurate for bounded scenes where a central object is the main part of the scene and no background information is present. However, the approximation deteriorates for highly elongated frustums. To avoid this, mip-NeRF 360 \cite{barron2022mip} instead uses a contracted space where points that are beyond the unit sphere are mapped using the function:
\begin{equation}
    \label{eq:contracted-space}
    f(\mathbf{x}) = \begin{cases}
        \mathbf{x} & \lVert \mathbf{x} \rVert \leq 1 \\
         \left( 2 - \frac{1}{\lVert \mathbf{x} \rVert}\right) \frac{\mathbf{x}}{\lVert \mathbf{x} \rVert} & \lVert \mathbf{x} \rVert > 1
    \end{cases}\,.
\end{equation}

\noindent
Subsequently, the new $\boldsymbol{\mu}$ and $\boldsymbol{\Sigma}$ values are given by $f(\boldsymbol{\mu})$ and $\mathbf{J}_f(\boldsymbol{\mu}) \boldsymbol{\Sigma} \mathbf{J}_f (\boldsymbol{\mu})^\top$, where $\mathbf{J}_f$ is the Jacobian matrix of $f$. Empirically, this re-parameterization now allows learning the representation of scenes with distant backgrounds (\ie over a longer depth of field). 

\section{{\name}}
\label{sec:method}
\noindent
In this paper, we present {\name} as an exploration of how the IPE approximations of earlier work \cite{barron2021mip, barron2022mip} based on conic parameterization can be replaced with a square pyramid-based formulation in order to obtain an exact IPE $\gamma_E$, as shown in \cref{fig:frustums}. The motivation behind this formulation is to match the volumetric rendering with the pixel footprint, which in turn is a rectangle. 

\begin{figure}[!t]
\centering
\subfloat{\includegraphics[width=0.9\linewidth]{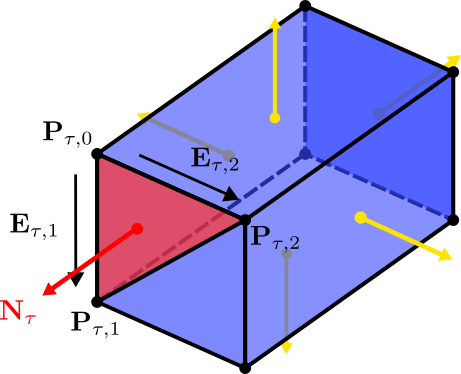}}
\hfil
\caption{Parameterization of triangular faces. The vertices are sorted counter-clockwise, so the normal vector to their plane points outside the frustum.}
\label{fig:triangles}
\end{figure}
\subsection{Volume of pyramidal frustums}
\noindent
A pyramidal frustum can be defined by a set of 8 vertices $\mathcal{V}=\left\{ \mathbf{v}_i \right\}_{i=1}^8$ and 6 quadrilateral faces $\mathcal{F}=\left\{f_j\right\}_{j=1}^6$. In order to get the volume in the denominator of \cref{eq:ipe}, we use the divergence theorem:
\begin{equation}
\label{eq:div_theorem}
    \iiint \nabla \cdot F dV = \oiint_{\partial S} F \cdot d \mathbf{S}\,,
\end{equation}
with $F = \frac{1}{3}\left[ x, y, z\right]^\top$, yielding to the solution for the volume as:
\begin{equation}
\label{eq:vol1}
    V = \iiint dV = \frac{1}{3} \oiint_{\partial S} \left[x, y, z\right] d \mathbf{S}\,.
\end{equation}
Without losing generality, we divide each face into triangles, giving a set of triangular faces $\mathcal{T}$ such that the polyhedra formed by faces $\mathcal{F}$ and $\mathcal{T}$ are the same. Each triangle $\tau$ is defined by three points $\mathbf{P}_{\tau, 0}, \mathbf{P}_{\tau, 1}$ and $\mathbf{P}_{\tau, 2}$, with $\mathbf{P}_{\tau, i} \in \mathcal{V}$, such that the cross product of the edges $\mathbf{E}_{\tau, 1} = \mathbf{P}_{\tau, 1} - \mathbf{P}_{\tau, 0}$ and $\mathbf{E}_{\tau, 2} = \mathbf{P}_{\tau, 2} - \mathbf{P}_{\tau, 0}$ points outside the frustum (\cref{fig:triangles}). As a result, \cref{eq:vol1} equates to the sum of the surface integral for each triangle $\tau \in \mathcal{T}$,
\begin{equation}
\label{eq:vol2}
    V = \frac{1}{3} \sum_{\tau \in \mathcal{T}} \iint_\tau \left[x, y, z\right] d \mathbf{S}\,.
\end{equation}
The points lying in the triangle $\triangle \mathbf{P}_{\tau, 0}\mathbf{P}_{\tau, 1}\mathbf{P}_{\tau, 2}$ can hence be parameterized as:
\begin{equation}
\label{eq:point_param}
\mathbf{P}_\tau(u, v) = \mathbf{P}_{\tau, 0} + u\mathbf{E}_{\tau, 1} + v \mathbf{E}_{\tau, 2}\,,
\end{equation}
\noindent
such that $0 \leq u \leq 1, 0 \leq v \leq 1$ and $u+v \leq 1$. The differential term of \cref{eq:vol2} is then:
\begin{align}
\label{eq:surf_diff}
   d \mathbf{S} &=  \left(\frac{\partial \mathbf{P}_\tau}{\partial u} \times \frac{\partial \mathbf{P}_\tau}{\partial v}\right) du dv\\
\label{eq:surf_diff2}
   d \mathbf{S} &=  \left(\mathbf{E}_{\tau, 1} \times \mathbf{E}_{\tau, 2}\right) du dv \triangleq \mathbf{N}_\tau du dv\,.
\end{align}
By substituting \cref{eq:surf_diff2} into \cref{eq:vol2}, and noting that $\left[x, y, z\right]=\mathbf{P}_\tau(u,v)$, we obtain:
\begin{equation}
\label{eq:vol3}
    V = \frac{1}{3} \sum_{\tau \in \mathcal{T}} \int_0^1 \int_0^{1-v} \mathbf{P}_\tau(u,v)^\top \mathbf{N}_\tau  du dv\,.
\end{equation}
Since the dot product of any point $\mathbf{P}_\tau$ in a face $\tau$ with a vector $\mathbf{N}_\tau$ normal to $\tau$ is constant, the product inside the integral of \cref{eq:vol3} is constant. Subsequently, $\mathbf{P}_\tau(u,v)$ can be replaced with any point, such as $\mathbf{P}_{\tau, 0}$. Finally, the required volume is obtained as:
\begin{align}
\label{eq:vol4}
\begin{split}
    V &= \frac{1}{3} \sum_{\tau \in \mathcal{T}}  \mathbf{P}_{\tau, 0}^\top \mathbf{N}_\tau  \int_0^1 \int_0^{1-v} du dv\\
    &= \frac{1}{6}  \sum_{\tau \in \mathcal{T}}  \mathbf{P}_{\tau, 0}^\top \mathbf{N}_\tau   \,.
\end{split}
\end{align}

\subsection{Integration over the PE Function}
\noindent
Following from earlier, we can obtain the numerator of  the IPE in \cref{eq:ipe} using the divergence theorem. We will base our analysis on the sine function and the $x$ coordinate, \ie, $\gamma(x) = \sin(2^l x)$. Substituting $F=\left[-\frac{1}{2^l} \cos(2^l x),0,0\right]^\top$ in \cref{eq:div_theorem} we obtain:
\begin{equation}
\label{eq:ipe-int1}
    \iiint \sin(2^l x)dV = \oiint_{\partial S} \left[-\frac{1}{2^l} \cos(2^l x), 0, 0 \right] d\mathbf{S}\,.
\end{equation}
Following the same strategy of dividing the surface into triangular faces as in the earlier volume calculation, \cref{eq:ipe-int1} can be written as:
\begin{equation}
\label{eq:ipe-int2}
    \iiint \sin(2^l x)dV =
    \sum_{\tau \in \mathcal{T}} \frac{1}{2^l} \sigma_{x,\tau} \mathbf{N}_\tau  \cdot \ihat  \,,
\end{equation}
where $\ihat$ is the unit vector in the $x$ direction and:
\begin{equation}
\label{eq:sigma-int2}
    \sigma_{x, \tau} = \int_0^1 \int_0^{1-v} -\cos(2^l x_\tau(u,v))du dv\,.
\end{equation}
From \cref{eq:point_param}, the $x$ coordinate can be parameterized as:
\begin{equation}
\label{eq:x_param}
    x_\tau (u,v) = x_{\tau, 0} + u (x_{\tau, 1} - x_{\tau, 0})+ v (x_{\tau, 2} - x_{\tau, 0})\,.
\end{equation}
\noindent
Substituting \cref{eq:x_param} in \cref{eq:sigma-int2} and solving the integral, we obtain:
\begin{equation}
\label{eq:sigma-int3}
\begin{split}
     \sigma_{x, \tau} = \frac{1}{2^{2l}} &\left( \frac{\cos(2^l x_{\tau, 0})}{ ( x_{\tau, 0} - x_{\tau, 1})(x_{\tau, 0} - x_{\tau, 2})} \right. \\
     &\left. +  \frac{\cos(2^l x_{\tau, 1})}{ (x_{\tau, 1} - x_{\tau, 0})(x_{\tau, 1} - x_{\tau, 2})} \right. \\
     &\left.+ \frac{\cos(2^l x_{\tau, 2})}{ (x_{\tau, 2} - x_{\tau, 0})(x_{\tau, 2} - x_{\tau, 1})} \right)\,.
\end{split}
\end{equation}
Furthermore, \cref{eq:sigma-int3} can be written as:
\begin{equation}
    \label{eq:sigma-int4}
    \sigma_{x,\tau} = \frac{1}{2^{2l}} \frac{ \det \left( \begin{bmatrix} 
    \mathbf{1} & \boldsymbol{x}_{\tau} & \cos(2^l \boldsymbol{x}_{\tau})
    \end{bmatrix} \right)
    }{\det \left( \begin{bmatrix} 
    \mathbf{1} & 
    \boldsymbol{x}_{\tau} & 
    \boldsymbol{x}_{\tau}^{\circ 2}\\
    \end{bmatrix} \right)}\,,
\end{equation}
where $\mathbf{1} = \left[ 1, 1, 1 \right]^\top$, $\boldsymbol{x}_{\tau} = \left[ x_{\tau, 0}, x_{\tau, 1}, x_{\tau, 2} \right]^\top$ and $(\cdot)^{\circ n}$ is the element-wise power.

In general, we can also obtain the expression in \cref{eq:ipe-int2} for the $k$-th coordinate of $\mathbf{x}$ as:
\begin{equation}
\label{eq:ipe-general-sin}
    \iiint \sin(2^l \mathbf{x}_k)dV = 
    \frac{1}{2^{3l}}\sum_{\tau \in \mathcal{T}}{ \sigma_{k,\tau}
    \mathbf{N}_\tau \cdot  \mathbf{e}_k} \,,
\end{equation}
\begin{equation}
 \sigma_{k,\tau} = \frac{ \det \left( \begin{bmatrix} 
    \mathbf{1} & \mathbf{X}_{\tau}^\top \mathbf{e}_k & \cos(2^l \mathbf{X}_{\tau}^\top \mathbf{e}_k)
    \end{bmatrix} \right)} {\det \left( \begin{bmatrix} 
    \mathbf{1} & 
    \mathbf{X}_{\tau}^\top \mathbf{e}_k & 
    (\mathbf{X}_{\tau}^\top \mathbf{e}_k)^{\circ 2}\\
    \end{bmatrix} \right)}\,,
\end{equation}
where $\mathbf{X}_\tau = \begin{bmatrix} \mathbf{P}_{\tau, 0} & \mathbf{P}_{\tau, 1} & \mathbf{P}_{\tau, 2} \end{bmatrix}$ and $\mathbf{e}_k$ are the vectors that form the canonical basis in $\mathbb{R}^3$. Similarly, the integral over the cosine function is defined as:
\begin{equation}
\label{eq:ipe-general-cos}
    \iiint \cos(2^l \mathbf{x}_k)dV = 
    \frac{1}{2^{3l}}\sum_{\tau \in \mathcal{T}}{ \xi_{k,\tau}
    \mathbf{N}_\tau \cdot \mathbf{e}_k} \,,
\end{equation}
where:
\begin{equation}
\label{eq:xi}
 \xi_{k,\tau} = -\frac{ \det \left( \begin{bmatrix} 
    \mathbf{1} & \mathbf{X}_{\tau}^\top \mathbf{e}_k & \sin(2^l \mathbf{X}_{\tau}^\top \mathbf{e}_k)
    \end{bmatrix} \right)} {\det \left( \begin{bmatrix} 
    \mathbf{1} & 
    \mathbf{X}_{\tau}^\top \mathbf{e}_k & 
    (\mathbf{X}_{\tau}^\top \mathbf{e}_k)^{\circ 2}\\
    \end{bmatrix} \right)}\,.
\end{equation}
Finally, we get the exact IPE (EIPE) of the frustum used by our {\name} approach by dividing \cref{eq:ipe-general-sin,eq:ipe-general-cos} by \cref{eq:vol4} as follows:
\begin{equation}
    \label{eq:exact-ipe}
    \gamma_E (\mathbf{x}, l; \mathcal{V}) = \frac{6}{2^{3l}}
    \begin{bmatrix}
        \frac{\sum_{\tau \in \mathcal{T}} \boldsymbol{\sigma}_\tau \circ \mathbf{N}_\tau}{\sum_{\tau \in \mathcal{T}} \mathbf{P}_{\tau, 0}^\top \mathbf{N}_\tau} \\
        \frac{\sum_{\tau \in \mathcal{T}} \boldsymbol{\xi}_\tau \circ \mathbf{N}_\tau}{\sum_{\tau \in \mathcal{T}} \mathbf{P}_{\tau, 0}^\top \mathbf{N}_\tau}\,,
    \end{bmatrix}
\end{equation}
\noindent
where $\boldsymbol{\sigma}_\tau = \begin{bmatrix}\sigma_{1, \tau} & \sigma_{2, \tau} & \sigma_{3, \tau} \end{bmatrix}^\top$ and $\boldsymbol{\xi}_\tau = \begin{bmatrix}\xi_{1, \tau} & \xi_{2, \tau} & \xi_{3, \tau} \end{bmatrix}^\top$. It's worth mentioning that \cref{eq:exact-ipe} fails when a coordinate value repeats in any of the points of a triangle (\ie, there is a triangle $\tau$ such that $\mathbf{P}_{\tau, i} = \mathbf{P}_{\tau, j}$ for a $i \neq j$). For these cases, \textit{l'Hopital's rule} can be used to evaluate this limit (see Supplementary Material).

\begin{figure*}[!t]
\centering
\subfloat{\includegraphics[width=\linewidth]{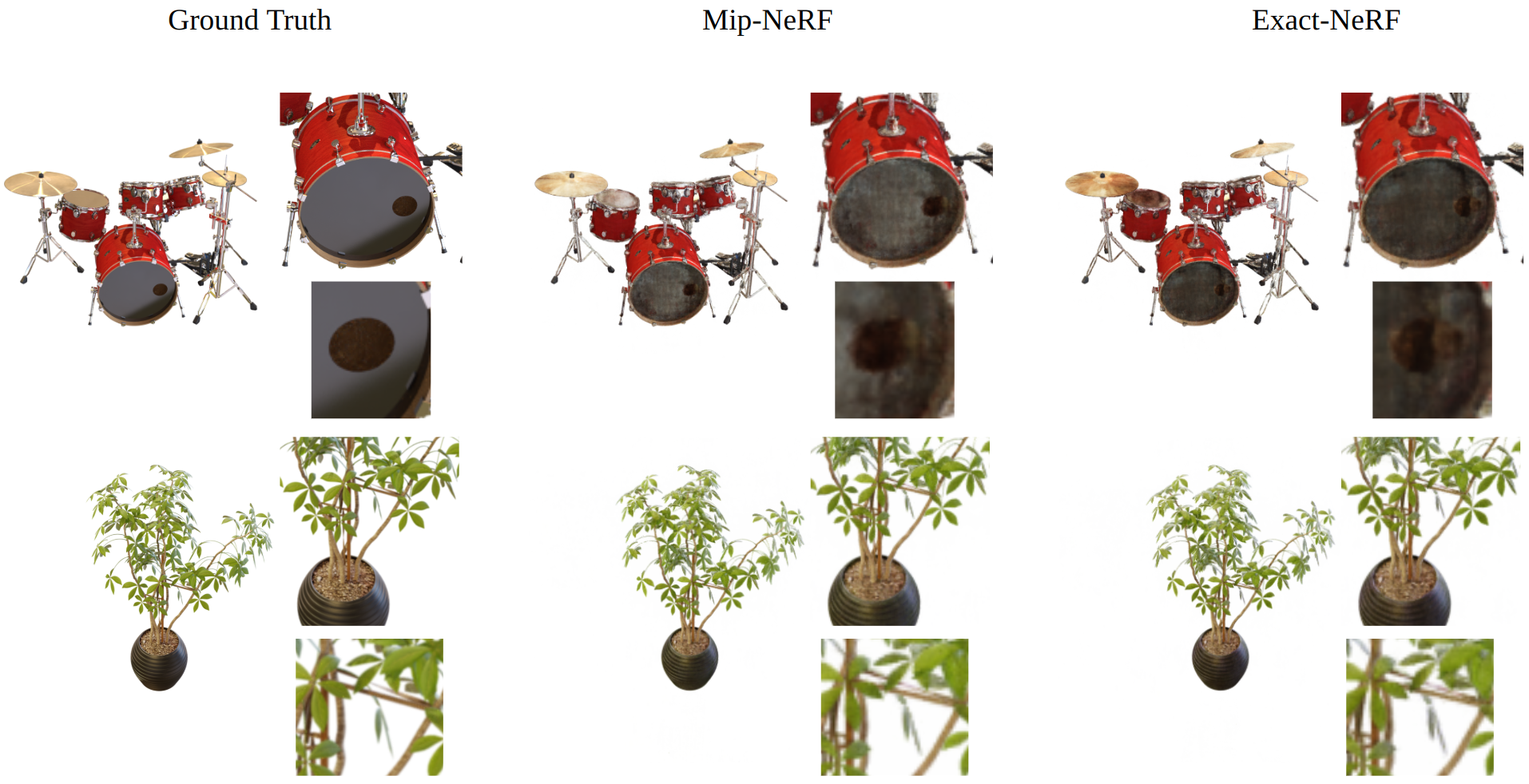}}
\hfil
\caption{Qualitative comparison between mip-NeRF and {\name} (ours) for the blender dataset. Our method matches the mip-NeRF rendering capability but also produces slightly sharper renderings (see the bass drum hole and the back leaves of the ficus).}
\label{fig:blender-results}
\end{figure*}

Despite starting our analysis with squared pyramids, it can be noted that \cref{eq:exact-ipe} is true for any set of vertices $\mathcal{V}$, meaning that this parameterization can be applied for any shape with known vertices. This is particularly useful for scenarios where the space may be deformed and frustums may not be perfect pyramids, such as in mip-Nerf 360 \cite{barron2022mip}. Additionally, it can be noted that our EIPE is multiplied by a factor of $2^{-3l}$, meaning that when $L \rightarrow \infty$ then $\gamma_E \rightarrow 0$ which hence makes our implementation robust to large values of $L$. This property of our {\name} formulation is consistent with that of the original mip-NeRF \cite{barron2021mip}.

\section{Implementation Details} 
\noindent
{\name} is implemented using the original code of mip-NeRF, which is based on JAXNeRF \cite{jaxnerf2020github}. Apart from the change of the positional encoding, no further modification is made. We use the same sampling strategy of ray intervals defined in \cref{eq:sampling_t}, but sampling $N+1$ points to define $N$ intervals. In order to obtain the vertices of the pyramid frustums, we use the coordinates of the corners of each pixel and multiply them by the $t_i$ values to get the front and back faces of the frustums. Double precision (64-bit float) is used for calculating the EIPE itself, as it relies upon arithmetic over very low numerical decimals that are otherwise prone to numerical precision error (see \cref{eq:sigma-int3}). After calculation, the EIPE result is transformed back to single precision (32-bit float).

We compare our implementation of {\name} against the original mip-NeRF baseline on the benchmark Blender dataset \cite{mildenhall2021nerf}, down-sampled by a factor of 2. We follow a similar training strategy as in mip-NeRF: training both models for 800k iterations (instead of 1 million, as we observed convergence at this point) with a batch size of 4096 using Adam optimization \cite{kingma2015adam} with a logarithmically annealed learning rate, $5 \times 10^{-4} \rightarrow 5 \times 10^{-6}$. All training is carried out using 2 $\times$ NVIDIA Tesla V100 GPU per scene.

Additionally, we compare the use of the EIPE against mip-NeRF 360 on the dataset of Barron \etal \cite{barron2022mip}. Similarly, we used the reference code from MultiNeRF \cite{multinerf2022}, which contains an implementation of mip-NeRF 360 \cite{barron2022mip},
RefNeRF \cite{verbin2022refnerf} and RawNeRF \cite{mildenhall2022rawnerf}. Pyramidal frustum vertices are contracted using \cref{eq:contracted-space} and the EIPE is obtained using the \cref{eq:exact-ipe} with the mapped vertices. We trained using a batch size of 8192 for 500k iterations using 4 $\times$ NVIDIA Tesla V100 GPU per scene. Aside from the use of the EIPE, all other settings remained unchanged from mip-NeRF 360 \cite{barron2022mip}.

\section{Results}
\label{sec:results}
\noindent
Mean PSNR, SSIM and LPIPS \cite{zhang2018perceptual} metrics are reported for our {\name} approach, mip-NeRF \cite{barron2021mip} and mip-NeRF 360 \cite{barron2022mip}. Additionally, we also report the DISTS \cite{ding2020image} metric since it provides another perceptual quality measurement. Similar to mip-NeRF, we also report an average metric: the geometric mean of the MSE = $10^{-\text{PSNR}/10}$, $\sqrt{1-\text{SSIM}}$, the LPIPS and the DISTS.

\begin{table}[t]
    \centering
    \resizebox{\linewidth}{!}{%
    \begin{tabular}{|c | c  c  c c  c |}
    \hline
    Model & PSNR $\uparrow$ & SSIM $\uparrow$ & LPIPS $\downarrow$& DISTS $\downarrow$& Avg $\downarrow$\\
    \hline
    Mip-NeRF & \textbf{34.766} & \textbf{0.9706} & 0.0675 & 0.0878 & \textbf{0.0242} \\
    {\name} (ours) & 34.707 & 0.9705 & \textbf{0.0667} & \textbf{0.0822} & \textbf{0.0242}\\
    \hline
    \end{tabular}%
    }
    
    \caption{Quantitative results comparing mip-NeRF and {\name} performance on the Blender dataset.}
    \label{Table:results-mipnerf}
\end{table}

\noindent
\textbf{Blender dataset:} In \cref{Table:results-mipnerf} we present a quantitative comparison between {\name} and mip-NeRF. It can be observed that our method matches the reconstruction performance of mip-NeRF, with a marginal decrease of the PSNR and SSIM and an increment in the LPIPS and DISTS metrics, but with identical average performance. This small decrement in the PSNR and SSIM metrics can be explained by the loss of precision in the calculation of small quantities involved in the EIPE. Alternative formulations using the same idea could be used (see Supplementary Material), but the intention of {\name} is to create a general approach for any volumetric positional encoding using the vertices of the volumetric region. \cref{fig:blender-results} shows a qualitative comparison between mip-NeRF and {\name}. It can be observed that {\name} is able to match the reconstruction performance of mip-NeRF. A closer examination reveals that {\name} creates sharper reconstructions in some regions, such as the hole in the bass drum or the leaves in the ficus, which is explained by mip-NeRF approximating the conical frustums as Gaussians. This is consistent with the increase in the LPIPS and DISTS, which are the perceptual similarity metrics. 
\begin{table}[t]
    \centering
    \resizebox{\linewidth}{!}{%
    \begin{tabular}{|c | c  c c  c  c |}
    \hline
    Model & PSNR $\uparrow$ & SSIM $\uparrow$ & LPIPS $\downarrow$& DISTS $\downarrow$& Avg $\downarrow$\\
    \hline
    Mip-NeRF 360 & \textbf{27.325} & \textbf{0.7942} & \textbf{0.6559} & \textbf{0.2438} & \textbf{0.1077} \\
    {\name} (ours) & 27.230 & 0.7881 & 0.6569 & 0.2452 & 0.1088 \\
    \hline
    \end{tabular}%
    }
    
    \caption{Comparison of the performance of {\name} with mip-NeRF 360 on the unbounded dataset of Barron \etal \cite{barron2022mip}.}
    \label{Table:results-mipnerf360}
\end{table}

\begin{figure*}[!t]
\centering
\subfloat{\includegraphics[width=\linewidth]{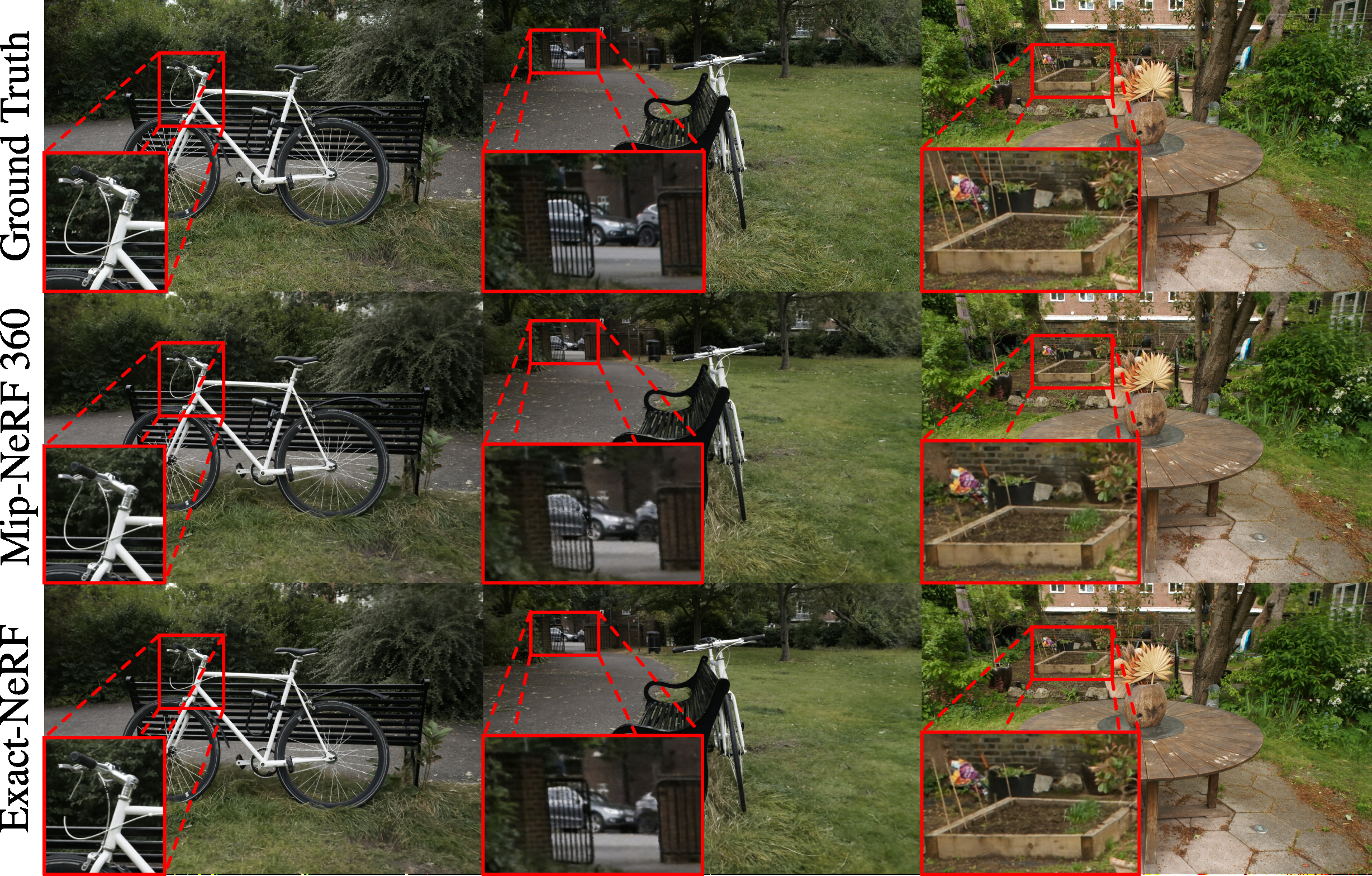}}
\hfil
\caption{Qualitative comparison between mip-NeRF 360 and {\name} (ours). (a) Our model, similar to mip-NeRF, struggles with tiny vessels. (b) {\name} shows cleaner renderings and (c) higher quality background reconstruction.}
\label{fig:bicycle-results}
\end{figure*}

\begin{figure*}[tbp]
\centering
\subfloat{\includegraphics[width=0.95\linewidth]{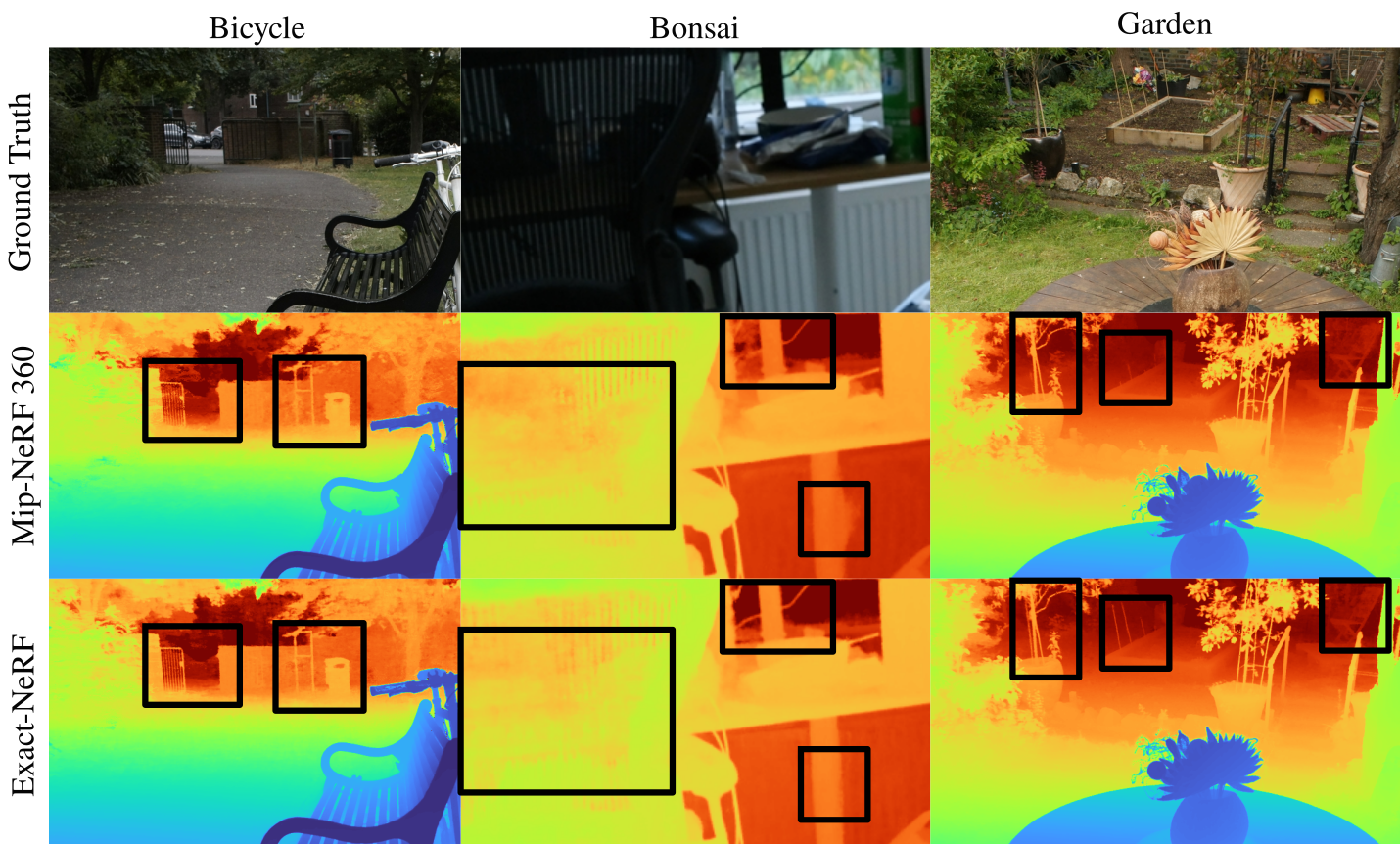}}
\hfil
\caption{Depth estimation for mip-NerF 360 and {\name}. Our approach shows better depth estimations for background regions (highlighted in the black boxes), although some artifacts in form of straight lines may appear, which is inherent in our pyramidal shapes.}
\label{fig:depth-results}
\end{figure*}

\noindent
\textbf{Mip-NeRF 360 dataset:} \cref{Table:results-mipnerf360} shows the results for the unbounded mip-NeRF 360 dataset. Despite {\name} having marginally weaker reconstruction metrics, it shows a competitive performance without any changes to the implementation of the EIPE used earlier with the bounded blender dataset, \ie, the contracted vertices were directly used without any further simplification or linearization, as in mip-NeRF 360 \cite{barron2022mip}. Similar to the blender dataset results, this decrement can be explained with the loss of precision, which suggests that an alternative implementation of \cref{eq:exact-ipe} may be needed. A qualitative comparison is shown in \cref{fig:bicycle-results}. It can be observed that tiny vessels are more problematic for {\name} (\cref{fig:bicycle-results}a), which can be explained again by the loss of precision. However, it is noted in \cref{fig:bicycle-results}b that the reconstruction of far regions in mip-NeRF 360 is noisier than {\name} (see  \cref{fig:bicycle-results}b,  grill and the car), which is a consequence of the poor approximation of the Gaussian region for far depth of field objects in the scene. \cref{fig:bicycle-results}c reveals another example of a clearer region in the {\name} reconstruction for the background detail. \cref{fig:depth-results} shows snapshots of the depth estimation for the bicycle, bonsai and garden scenes. Consistent with the colour reconstructions, some background regions have a more detailed estimation. It is also noticed (not shown) that despite {\name} having a smoother depth estimation, it may show some artifacts in the form of straight lines, which may be caused by the shape of the pyramidal frustums. It is worth reminding that our implementation of the EIPE in mip-NeRF 360 is identical to the EIPE in mip-NeRF.

\begin{figure}[t]
\centering
\subfloat{\includegraphics[width=0.95\linewidth]{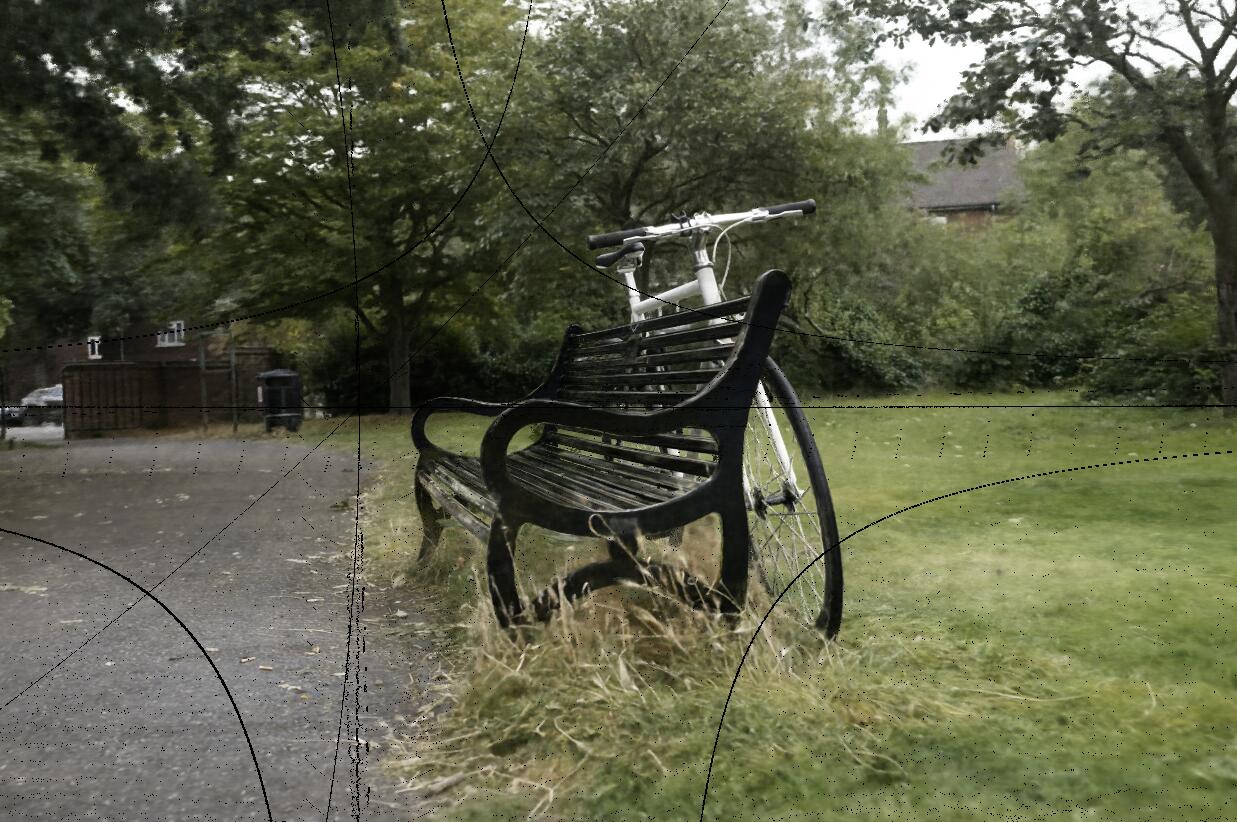}}
\caption{Numerical underflow artifacts in {\name}.}
\label{fig:underflow-artifacts}
\end{figure}

\noindent
\textbf{Impact of Numerical Underflow} As seen in \cref{sec:method}, {\name} may suffer from numerical underflow when the difference of a component of two points $\Delta = x_{\tau, i} - x_{\tau, j}$ is too close to zero ($\Delta \rightarrow 0$). In the case of this difference being precisely zero, the limit can be found using \textit{l'Hopital's rule}, as is further developed in Appendix A.1. However, if this value is not zero but approximately zero, numerical underflow could lead to exploding values in \cref{eq:sigma-int3}. This error hinders the training of the MLP since the IPE is bounded to the interval $\left[ -1, 1\right]$ by definition (\cref{eq:ipe}). An example of the effect of numerical underflow in our method applied under the mip-NeRF 360 framework is shown in \cref{fig:underflow-artifacts}. The black lines are the location of such instances where underflow occurs. The curvature of these lines is a direct consequence of the contracted space used in mip-NeRF 360. In order to eliminate this effect, we use double precision for the calculation of the EIPE. Additionally, all differences of a coordinate which are less than $1\times10^{-6}$ are set to zero and reformulated using  \textit{l'Hopital's rule}.

\section{Conclusion}
\noindent
In this work, we present {\name}, a novel precise volumetric parameterization for neural radiance fields (NeRF). In contrast to conical frustum approximation via a multivariate Gaussian in mip-NeRF \cite{barron2021mip}, {\name} uses a novel pyramidal parameterization to encode 3D regions using an Exact Integrated Positional Encoding (EIPE). The EIPE applies the divergence theorem to compute the exact value of the positional encoding (an array of sine and cosines) in a pyramidal frustum using the coordinates of the vertices that define the region. Our proposed EIPE methodology can be applied to any such architecture that performs volumetric positional encoding from simple knowledge of the pyramidal frustum vertices without the need for further processing.

We compare {\name} against mip-NeRF on the blender dataset, showing a matching performance with a marginal decrease in PSNR and SSIM but an overall improvement in the perceptual metric, LPIPS. Qualitatively our approach exhibits slightly cleaner and sharper reconstructions of edges than mip-NeRF \cite{barron2021mip}. 

We similarly compare {\name} with mip-NeRF 360 \cite{barron2022mip}. Despite {\name} showing a marginal decrease in performance metrics, it illustrates the capability of the EIPE on a different architecture without further modification. {\name} obtains sharper renderings of distant (far depth of field) regions and areas where mip-NeRF 360 presents some noise, but it fails to reconstruct tiny vessels in near regions. The qualitative depth estimations maps also confirm these results. The marginal decrease in performance of our {\name} method can be attributed to numerical underflow and some artifacts caused by the choice of a step-function-based square pyramidal parameterization. In addition, our results suggest using a combined encoding such that the EIPE is used for distance objects, where it is more stable and accurate. Although alternative solutions can be obtained by restricting the analysis to rectangular pyramids, our aim is to introduce a general framework that can be applied to any representation of a 3D region with known vertices. The investigation of more stable representations and the performance of different shapes for modelling 3D regions under a neural rendering context remains an area for future work.

\section*{Acknowledgments}
\noindent
This work is partially supported by the Mexican Council of Science and Technology (CONACyT). 

{\small
\bibliographystyle{ieee_fullname}
\bibliography{egbib}

\begin{thebibliography}{10}\itemsep=-1pt

\bibitem{barron2021mip}
Jonathan~T Barron, Ben Mildenhall, Matthew Tancik, Peter Hedman, Ricardo
  Martin-Brualla, and Pratul~P Srinivasan.
\newblock Mip-nerf: A multiscale representation for anti-aliasing neural
  radiance fields.
\newblock In {\em Proceedings of the IEEE Conf. Comput. Vis. Pattern Recog.},
  pages 5855--5864, 2021.

\bibitem{barron2022mip}
Jonathan~T Barron, Ben Mildenhall, Dor Verbin, Pratul~P Srinivasan, and Peter
  Hedman.
\newblock Mip-nerf 360: Unbounded anti-aliased neural radiance fields.
\newblock In {\em Proceedings of the IEEE Conf. Comput. Vis. Pattern Recog.},
  pages 5470--5479, 2022.

\bibitem{coronafigueroa2022mednerf}
Abril Corona-Figueroa, Jonathan Frawley, Sam~Bond Taylor, Sarath Bethapudi,
  Hubert P.~H. Shum, and Chris~G. Willcocks.
\newblock Mednerf: Medical neural radiance fields for reconstructing 3d-aware
  ct-projections from a single x-ray.
\newblock In {\em Proceedings of the 2022 44th Annual International Conference
  of the IEEE Engineering in Medicine \& Biology Society}, pages 3843--3848,
  2022.

\bibitem{jaxnerf2020github}
Boyang Deng, Jonathan~T. Barron, and Pratul~P. Srinivasan.
\newblock {JaxNeRF}: an efficient {JAX} implementation of {NeRF}.
\newblock
  \url{https://github.com/google-research/google-research/tree/master/jaxnerf},
  2020.

\bibitem{Deng_2022_CVPR}
Kangle Deng, Andrew Liu, Jun-Yan Zhu, and Deva Ramanan.
\newblock Depth-supervised nerf: Fewer views and faster training for free.
\newblock In {\em Proceedings of the IEEE Conf. Comput. Vis. Pattern Recog.},
  pages 12882--12891, June 2022.

\bibitem{mipnerfrgbd}
Arnab Dey, Yassine Ahmine, and Andrew~I. Comport.
\newblock Mip-{NeRF} {RGB}-d: Depth assisted fast neural radiance fields.
\newblock {\em Journal of {WSCG}}, 30(1-2):34--43, 2022.

\bibitem{ding2020image}
Keyan Ding, Kede Ma, Shiqi Wang, and Eero~P Simoncelli.
\newblock Image quality assessment: Unifying structure and texture similarity.
\newblock {\em Proceedingf the IEEE Trans. Pattern Anal. Mach. Intell.}, 2020.

\bibitem{ding2021comparison}
Keyan Ding, Kede Ma, Shiqi Wang, and Eero~P Simoncelli.
\newblock Comparison of full-reference image quality models for optimization of
  image processing systems.
\newblock {\em Int. J. Comp. Vis.}, 129(4):1258--1281, 2021.

\bibitem{Fridovich-Keil_2022_CVPR}
Sara Fridovich-Keil, Alex Yu, Matthew Tancik, Qinhong Chen, Benjamin Recht, and
  Angjoo Kanazawa.
\newblock Plenoxels: Radiance fields without neural networks.
\newblock In {\em Proceedings of the IEEE Conf. Comput. Vis. Pattern Recog.},
  pages 5501--5510, June 2022.

\bibitem{guan2022neurofluid}
Shanyan Guan, Huayu Deng, Yunbo Wang, and Xiaokang Yang.
\newblock Neurofluid: Fluid dynamics grounding with particle-driven neural
  radiance fields.
\newblock In {\em Proceedings of the Int. Conf. on Mach. Learning}, 2022.

\bibitem{Hedman_2021_ICCV}
Peter Hedman, Pratul~P. Srinivasan, Ben Mildenhall, Jonathan~T. Barron, and
  Paul Debevec.
\newblock Baking neural radiance fields for real-time view synthesis.
\newblock In {\em Proceedings of the Int. Conf. Comput. Vis.}, pages
  5875--5884, October 2021.

\bibitem{hore2010}
Alain Horé and Djemel Ziou.
\newblock Image quality metrics: Psnr vs. ssim.
\newblock In {\em 2010 20th Int. Conf. on Pattern Recog.}, pages 2366--2369,
  2010.

\bibitem{Hu_2022_CVPR}
Tao Hu, Shu Liu, Yilun Chen, Tiancheng Shen, and Jiaya Jia.
\newblock Efficientnerf efficient neural radiance fields.
\newblock In {\em Proceedings of the IEEE Conf. Comput. Vis. Pattern Recog.},
  pages 12902--12911, June 2022.

\bibitem{huang2022hdr}
Xin Huang, Qi Zhang, Ying Feng, Hongdong Li, Xuan Wang, and Qing Wang.
\newblock Hdr-nerf: High dynamic range neural radiance fields.
\newblock In {\em Proceedings of the IEEE Conf. Comput. Vis. Pattern Recog.},
  pages 18398--18408, 2022.

\bibitem{kingma2015adam}
Diederik~P Kingma and Jimmy Ba.
\newblock Adam: A method for stochastic optimization.
\newblock In {\em Proceedings of the Int. Conf. Learn. Represent.}, 2015.

\bibitem{kotevski2009}
Zoran Kotevski and Pece Mitrevski.
\newblock Experimental comparison of psnr and ssim metrics for video quality
  estimation.
\newblock In Danco Davcev and Jorge~Marx G{\'o}mez, editors, {\em ICT
  Innovations 2009}, pages 357--366, Berlin, Heidelberg, 2010. Springer Berlin
  Heidelberg.

\bibitem{levis2022gravitationally}
Aviad Levis, Pratul~P Srinivasan, Andrew~A Chael, Ren Ng, and Katherine~L
  Bouman.
\newblock Gravitationally lensed black hole emission tomography.
\newblock In {\em Proceedings of the IEEE Conf. Comput. Vis. Pattern Recog.},
  pages 19841--19850, 2022.

\bibitem{lin2021barf}
Chen-Hsuan Lin, Wei-Chiu Ma, Antonio Torralba, and Simon Lucey.
\newblock Barf: Bundle-adjusting neural radiance fields.
\newblock In {\em Proceedings of the Int. Conf. Comput. Vis.}, 2021.

\bibitem{SMPL:2015}
Matthew Loper, Naureen Mahmood, Javier Romero, Gerard Pons-Moll, and Michael~J.
  Black.
\newblock {SMPL}: A skinned multi-person linear model.
\newblock {\em Proceedings of the ACM Trans. Graph.}, 34(6):248:1--248:16, Oct.
  2015.

\bibitem{ma2022deblur}
Li Ma, Xiaoyu Li, Jing Liao, Qi Zhang, Xuan Wang, Jue Wang, and Pedro~V Sander.
\newblock Deblur-nerf: Neural radiance fields from blurry images.
\newblock In {\em Proceedings of the IEEE Conf. Comput. Vis. Pattern Recog.},
  pages 12861--12870, 2022.

\bibitem{martin2021nerf}
Ricardo Martin-Brualla, Noha Radwan, Mehdi~SM Sajjadi, Jonathan~T Barron,
  Alexey Dosovitskiy, and Daniel Duckworth.
\newblock Nerf in the wild: Neural radiance fields for unconstrained photo
  collections.
\newblock In {\em Proceedings of the IEEE Conf. Comput. Vis. Pattern Recog.},
  pages 7210--7219, 2021.

\bibitem{mildenhall2022rawnerf}
Ben Mildenhall, Peter Hedman, Ricardo Martin-Brualla, Pratul~P Srinivasan, and
  Jonathan~T Barron.
\newblock Nerf in the dark: High dynamic range view synthesis from noisy raw
  images.
\newblock In {\em Proceedings of the IEEE Conf. Comput. Vis. Pattern Recog.},
  pages 16190--16199, 2022.

\bibitem{mildenhall2021nerf}
Ben Mildenhall, Pratul~P Srinivasan, Matthew Tancik, Jonathan~T Barron, Ravi
  Ramamoorthi, and Ren Ng.
\newblock Nerf: Representing scenes as neural radiance fields for view
  synthesis.
\newblock {\em Communications of the ACM}, 65(1):99--106, 2021.

\bibitem{multinerf2022}
Ben Mildenhall, Dor Verbin, Pratul~P. Srinivasan, Peter Hedman, Ricardo
  Martin-Brualla, and Jonathan~T. Barron.
\newblock {MultiNeRF}: {A} {Code} {Release} for {Mip-NeRF} 360, {Ref-NeRF}, and
  {RawNeRF}.
\newblock \url{https://github.com/google-research/multinerf}, 2022.

\bibitem{ndajah2010ssim}
Peter Ndajah, Hisakazu Kikuchi, Masahiro Yukawa, Hidenori Watanabe, and Shogo
  Muramatsu.
\newblock Ssim image quality metric for denoised images.
\newblock In {\em Proceedings of the 3rd WSEAS Int. Conf. on Visualization,
  Imaging and Simulation}, pages 53--58, 2010.

\bibitem{neff2021donerf}
Thomas Neff, Pascal Stadlbauer, Mathias Parger, Andreas Kurz, Joerg~H. Mueller,
  Chakravarty R.~Alla Chaitanya, Anton~S. Kaplanyan, and Markus Steinberger.
\newblock {DONeRF: Towards Real-Time Rendering of Compact Neural Radiance
  Fields using Depth Oracle Networks}.
\newblock {\em Computer Graphics Forum}, 40(4), 2021.

\bibitem{Niemeyer2021Regnerf}
Michael Niemeyer, Jonathan~T. Barron, Ben Mildenhall, Mehdi S.~M. Sajjadi,
  Andreas Geiger, and Noha Radwan.
\newblock Regnerf: Regularizing neural radiance fields for view synthesis from
  sparse inputs.
\newblock In {\em Proceedings of the IEEE Conf. Comput. Vis. Pattern Recog.},
  2022.

\bibitem{spectralbias}
Nasim Rahaman, Aristide Baratin, Devansh Arpit, Felix Draxler, Min Lin, Fred
  Hamprecht, Yoshua Bengio, and Aaron Courville.
\newblock On the spectral bias of neural networks.
\newblock In {\em Proceedings of the Int. Conf. on Mach. Learning}, pages
  5301--5310. PMLR, 2019.

\bibitem{rematas2022urban}
Konstantinos Rematas, Andrew Liu, Pratul~P Srinivasan, Jonathan~T Barron,
  Andrea Tagliasacchi, Thomas Funkhouser, and Vittorio Ferrari.
\newblock Urban radiance fields.
\newblock In {\em Proceedings of the IEEE Conf. Comput. Vis. Pattern Recog.},
  pages 12932--12942, 2022.

\bibitem{sharif2018suitability}
Mahmood Sharif, Lujo Bauer, and Michael~K Reiter.
\newblock On the suitability of lp-norms for creating and preventing
  adversarial examples.
\newblock In {\em Proceedings of the IEEE Conf. Comput. Vis. Pattern Recog.
  Workshops}, pages 1605--1613, 2018.

\bibitem{sitzmann2021light}
Vincent Sitzmann, Semon Rezchikov, Bill Freeman, Josh Tenenbaum, and Fredo
  Durand.
\newblock Light field networks: Neural scene representations with
  single-evaluation rendering.
\newblock {\em Adv. Neural Inform. Process. Syst.}, 34:19313--19325, 2021.

\bibitem{tancik2022block}
Matthew Tancik, Vincent Casser, Xinchen Yan, Sabeek Pradhan, Ben Mildenhall,
  Pratul~P Srinivasan, Jonathan~T Barron, and Henrik Kretzschmar.
\newblock Block-nerf: Scalable large scene neural view synthesis.
\newblock In {\em Proceedings of the IEEE Conf. Comput. Vis. Pattern Recog.},
  pages 8248--8258, 2022.

\bibitem{verbin2022refnerf}
Dor Verbin, Peter Hedman, Ben Mildenhall, Todd Zickler, Jonathan~T. Barron, and
  Pratul~P. Srinivasan.
\newblock {Ref-NeRF}: Structured view-dependent appearance for neural radiance
  fields.
\newblock {\em Proceedings of the IEEE Conf. Comput. Vis. Pattern Recog.},
  2022.

\bibitem{wang2004}
Zhou Wang, A.C. Bovik, H.R. Sheikh, and E.P. Simoncelli.
\newblock Image quality assessment: from error visibility to structural
  similarity.
\newblock {\em IEEE Transactions on Image Processing}, 13(4):600--612, 2004.

\bibitem{mse-crit-1-2009}
Zhou Wang and Alan~C. Bovik.
\newblock Mean squared error: Love it or leave it? a new look at signal
  fidelity measures.
\newblock {\em IEEE Signal Processing Magazine}, 26(1):98--117, 2009.

\bibitem{xiangli2021citynerf}
Yuanbo Xiangli, Linning Xu, Xingang Pan, Nanxuan Zhao, Anyi Rao, Christian
  Theobalt, Bo Dai, and Dahua Lin.
\newblock Bungeenerf: Progressive neural radiance field for extreme multi-scale
  scene rendering.
\newblock In {\em Proceedings of the Eur. Conf. Comput. Vis.}, 2022.

\bibitem{recursive-nerf}
Guo-Wei Yang, Wen-Yang Zhou, Hao-Yang Peng, Dun Liang, Tai-Jiang Mu, and
  Shi-Min Hu.
\newblock Recursive-nerf: An efficient and dynamically growing nerf.
\newblock {\em IEEE Transactions on Visualization and Computer Graphics}, pages
  1--14, 2022.

\bibitem{Yu_2021_ICCV}
Alex Yu, Ruilong Li, Matthew Tancik, Hao Li, Ren Ng, and Angjoo Kanazawa.
\newblock Plenoctrees for real-time rendering of neural radiance fields.
\newblock In {\em Proceedings of the Int. Conf. Comput. Vis.}, pages
  5752--5761, October 2021.

\bibitem{zhang2020nerf++}
Kai Zhang, Gernot Riegler, Noah Snavely, and Vladlen Koltun.
\newblock Nerf++: Analyzing and improving neural radiance fields.
\newblock {\em arXiv preprint arXiv:2010.07492}, 2020.

\bibitem{zhang2018perceptual}
Richard Zhang, Phillip Isola, Alexei~A Efros, Eli Shechtman, and Oliver Wang.
\newblock The unreasonable effectiveness of deep features as a perceptual
  metric.
\newblock In {\em Proceedings of the IEEE Conf. Comput. Vis. Pattern Recog.},
  2018.

\end{thebibliography}
}
\clearpage
\appendix
\renewcommand\thesection{\Alph{section}}
\section{Additional Formulation}
\noindent
In this section, we will present some additional formulations used in our model  that do not affect the results presented in the paper. Notwithstanding that our proposed EIPE works for any polyhedron (which is the reason why it can be used under the mip-NeRF 360 \cite{barron2022mip} architecture without further treatment), we also present an alternative formulation of the EIPE for the particular case of strictly square pyramids. 

\subsection{Indeterminate cases for the EIPE}
\label{sec:indeterminateIPE}
\noindent
By simplifying \cref{eq:sigma-int3} we obtain:
\begin{equation}
    \label{eq:simp-sigma}
    \sigma_{x, \tau} = 
    \frac{
    \begin{aligned}[t]
        &(x_{\tau, 2} - x_{\tau, 1}) \cos(2^l x_{\tau, 0}) \\
        +&(x_{\tau, 0} - x_{\tau, 2}) \cos(2^l x_{\tau, 1}) \\
        +&(x_{\tau, 1} - x_{\tau, 0}) \cos(2^l x_{\tau, 2}) 
    \end{aligned}}{2^{2l} (x_{\tau, 1} - x_{\tau, 0}) (x_{\tau, 2} - x_{\tau, 0}) (x_{\tau, 2} - x_{\tau, 1})}\,.
\end{equation}
From \cref{eq:simp-sigma} we observe that an indetermination occurs for the case of two points in the triangle $\tau$ sharing the same coordinate, such that $x_{\tau, i} = x_{\tau, j}, i \neq j$. In order to get a valid value for these cases, we get the limit when those two coordinates approach. We can write \cref{eq:simp-sigma} as:
\begin{equation}
    \label{eq:simp-sigma-2}
    \sigma_{x, \tau} = 
    \frac{f(x_{\tau, 0}, x_{\tau, 1}, x_{\tau, 2})}{2^{2l} g(x_{\tau, 0}, x_{\tau, 1}, x_{\tau, 2})}\,.
\end{equation}
Then, we obtain the value for the case of $x_{\tau, 0} = x_{\tau, 1}$ using \textit{l'Hopital's rule}:
\begin{align}
&\lim_{x_{\tau, 0} \rightarrow x_{\tau, 1}}  \frac{f(x_{\tau, 0}, x_{\tau, 1}, x_{\tau, 2})}{2^{2l} g(x_{\tau, 0}, x_{\tau, 1}, x_{\tau, 2})} = \\
&\lim_{x_{\tau, 0} \rightarrow x_{\tau, 1}} \frac{ \frac{\partial}{\partial x_{\tau, 0}} f(x_{\tau, 0}, x_{\tau, 1}, x_{\tau, 2})}{2^{2l} \frac{\partial}{\partial x_{\tau, 0}} g(x_{\tau, 0}, x_{\tau, 1}, x_{\tau, 2})} = \\
\label{eq:hopitalx0}
&\lim_{x_{\tau, 0} \rightarrow x_{\tau, 1}} \frac{
\begin{pmatrix}
-2^l (x_{\tau, 2} - x_{\tau, 1}) \sin(2^l x_{\tau, 0})\\
+ \cos(2^l x_{\tau, 1})-\cos(2^l x_{\tau, 2})
\end{pmatrix}
}{2^{2l}(x_{\tau, 2} - x_{\tau, 1})(2x_{\tau, 0} - x_{\tau, 1} - x_{\tau, 2})} =
\end{align}
\begin{equation}
    \label{eq:x0tox1}
    \frac{2^l (x_{\tau, 2} - x_{\tau, 1}) \sin(2^l x_{\tau, 1})
- \cos(2^l x_{\tau, 1}) + \cos(2^l x_{\tau, 2})}{2^{2l}(x_{\tau, 2} - x_{\tau, 1})^2}\,.
\end{equation}
Similarly, from \cref{eq:hopitalx0}, we evaluate the case $x_{\tau, 0} = x_{\tau, 2}$:
\begin{equation}
    \label{eq:x0tox2}
    \frac{-2^l (x_{\tau, 2} - x_{\tau, 1}) \sin(2^l x_{\tau, 2})
+ \cos(2^l x_{\tau, 1})-\cos(2^l x_{\tau, 2})}{2^{2l}(x_{\tau, 2} - x_{\tau, 1})^2}\,.
\end{equation}
For the case when $x_{\tau, 1} = x_{\tau, 2}$, we differentiate with respect to  $x_{\tau, 1}$ to obtain the corresponding value:
\begin{align}
&\lim_{x_{\tau, 1} \rightarrow x_{\tau, 2}}  \frac{f(x_{\tau, 0}, x_{\tau, 1}, x_{\tau, 2})}{2^{2l} g(x_{\tau, 0}, x_{\tau, 1}, x_{\tau, 2})} = \\
\label{eq:hopitalx1}
&\lim_{x_{\tau, 1} \rightarrow x_{\tau, 2}} \frac{
\begin{pmatrix}
- \cos(2^l x_{\tau, 0}) +\cos(2^l x_{\tau, 2}) \\
+ 2^l (x_{\tau, 2} - x_{\tau, 0}) \sin(2^l x_{\tau, 1})
\end{pmatrix}
}{2^{2l}(x_{\tau, 2} - x_{\tau, 0})(x_{\tau, 0} + x_{\tau, 2} - 2x_{\tau, 1})} =
\end{align}
\begin{equation}
    \label{eq:x1tox2}
    \frac{-2^l (x_{\tau, 2} - x_{\tau, 0}) \sin(2^l x_{\tau, 1})
+ \cos(2^l x_{\tau, 0}) - \cos(2^l x_{\tau, 2})}{2^{2l}(x_{\tau, 2} - x_{\tau, 0})^2}\,.
\end{equation}
Finally, when $x_{\tau, 0} = x_{\tau, 1} = x_{\tau, 2}$, we use again the \textit{l'Hopital's rule} on \cref{eq:hopitalx0} and differentiate again with respect to $x_{\tau, 0}$ to obtain:
\begin{equation}
    \label{eq:x0tox1tox2}
    \lim_{x_{\tau, 0} \rightarrow x_{\tau, 1} \rightarrow x_{\tau, 2}} \sigma_{x, \tau} = - \frac{1}{2} \cos(2^l x_{\tau, 0}) \,.
\end{equation}
Using the same approach, we can find the following expressions for $\xi_{x, \tau}$ (\cref{eq:xi}):
\begin{equation}
        \lim_{x_{\tau,0} \rightarrow x_{\tau, 1}} \xi_{x, \tau} = 
        \frac{
        \begin{pmatrix}
            2^l (x_{\tau, 2} - x_{\tau, 1}) \cos(2^l x_{\tau, 1}) \\
           + \sin(2^l x_{\tau, 1}) - \sin(2^l x_{\tau, 2})
        \end{pmatrix}
        }{2^{2l}(x_{\tau, 2} - x_{\tau, 1})^2}
\end{equation}
\begin{equation}
        \lim_{x_{\tau,0} \rightarrow x_{\tau, 2}} \xi_{x, \tau} = \frac{
        \begin{pmatrix}
            -2^l (x_{\tau, 2} - x_{\tau, 1}) \cos(2^l x_{\tau, 2}) \\
           - \sin(2^l x_{\tau, 1}) + \sin(2^l x_{\tau, 2})
        \end{pmatrix}
        }{2^{2l}(x_{\tau, 2} - x_{\tau, 1})^2}
\end{equation}
\begin{equation}
        \lim_{x_{\tau,1} \rightarrow x_{\tau, 2}} \xi_{x, \tau} = \frac{
        \begin{pmatrix}
            -2^l (x_{\tau, 2} - x_{\tau, 0}) \cos(2^l x_{\tau, 2})\\
           - \sin(2^l x_{\tau, 0}) + \sin(2^l x_{\tau, 2})
        \end{pmatrix}
        }{2^{2l}(x_{\tau, 2} - x_{\tau, 0})^2}
\end{equation}
\begin{equation}
    \label{eq:x0tox1tox2_xi}
    \lim_{x_{\tau, 0} \rightarrow x_{\tau, 1} \rightarrow x_{\tau, 2}} \xi_{x, \tau} = \frac{1}{2} \sin(2^l x_{\tau, 0}) \,.
\end{equation}
Similar expressions can be obtained for the $y$ and $z$ coordinates.
\begin{figure}[t]
\centering
\subfloat{\includegraphics[width=\linewidth]{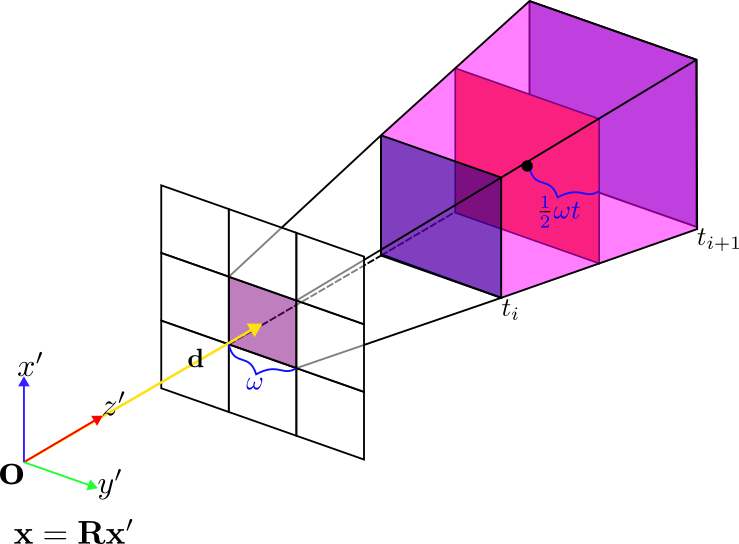}}
\caption{Parameterization of the square pyramid using the pixel width $\omega$.}
\label{fig:width-parameterization}
\end{figure}

\subsection{Alternative EIPE for Squared Pyramids}
\noindent
As mentioned earlier, our EIPE in \cref{eq:exact-ipe} can be used for any shape whose vertices are known. However, the computational cost increases if the 3D shape is complex since a larger number of triangular faces will need to be processed. For more efficient methods, we can focus our analysis on specific shapes. Particular to our scenario, we can obtain an alternative EIPE exclusively for a square pyramid (note that this will not be the case for the contraction function in mip-NeRF 360) with a known camera pose $[ \mathbf{R} | \mathbf{o}]$ and pixel width $\omega$ (similar to $\dot{r}$ in mip-NeRF). From \cref{fig:width-parameterization}, we calculate the volume of the frustum as
\begin{align}
    V&= \int_{t_i}^{t_{i+1}} \int_{-\omega z /2}^{\omega z /2} \int_{-\omega z /2}^{\omega z /2} dx'dy'dz' \\
    V&= \frac{\omega^2}{3} \left( t_{i+1}^3 - t_i^3\right)\,.
\end{align}
The numerator in \cref{eq:ipe} for the $x$ coordinate can be obtained in the same way:
\begin{align}
\label{eq:num-int}
    I_x = \int_{t_i}^{t_{i+1}} \int_{-\omega z /2}^{\omega z /2} \int_{-\omega z /2}^{\omega z /2} \sin(2^l x) dx'dy'dz'\,.
\end{align}
Since the camera pose is known, we can express $x$ as 
\begin{equation}
\label{eq:change-var}
    x = r_{11} x' + r_{12} y' + r_{13} z' + o_1\,,
\end{equation}
where $r_{ij}$ is an element of the rotation matrix $\mathbf{R}$ and $o_1$ is the first element of $\mathbf{o}$. Substituting \cref{eq:change-var} in \cref{eq:num-int} (and omitting the integration limits for clarity):
\begin{equation}
\label{eq:num-int-var}
I_x = \iiint \sin(2^l (r_{11} x' + r_{12} y' + r_{13} z' + o_1)) dx'dy'dz'\,.
\end{equation}
The solution to the integral in \cref{eq:num-int-var} is then:
\begin{align}
    \label{eq:pyramid-ipe}
    I_x &= \frac{1}{2^{3l} r_{11} r_{12}} \left[ \frac{C_1}{\zeta_1} -  \frac{C_2}{\zeta_2} -  \frac{C_3}{\zeta_3} +  \frac{C_4}{\zeta_4}\right]\,,\\
    C_j &= \cos\left(2^l(t_{i+1} \zeta_j+o_1)\right) - \cos\left(2^l(t_i \zeta_j+o_1)\right)\,,\\
    \zeta_j &= \boldsymbol{\eta}_j^\top \begin{bmatrix}r_{11}\\r_{12}\\r_{13}\end{bmatrix}\,,\\
    \boldsymbol{\eta}_1 &= \begin{bmatrix}\frac{\omega}{2}\\\frac{\omega}{2}\\1\end{bmatrix}\!,\boldsymbol{\eta}_2 = \begin{bmatrix}-\frac{\omega}{2}\\\frac{\omega}{2}\\1\end{bmatrix}\!,\boldsymbol{\eta}_3 = \begin{bmatrix}\frac{\omega}{2}\\-\frac{\omega}{2}\\1\end{bmatrix}\!,\boldsymbol{\eta}_4 = \begin{bmatrix}-\frac{\omega}{2}\\-\frac{\omega}{2}\\1\end{bmatrix}\,.
\end{align}
Similarly to the EIPE in \cref{eq:exact-ipe}, an indeterminate value arises in \cref{eq:pyramid-ipe} for $r_{11} = 0$ and $r_{12} = 0$. For these cases, \textit{l'Hopital's rule} can be used as in \cref{sec:indeterminateIPE} or \cref{eq:num-int-var} can be solved by substituting $r_{11} = 0$ and $r_{12} = 0$. We omit these calculations for brevity.

\begin{figure*}[h!]
\captionsetup[subfigure]{aboveskip=-12pt,belowskip=5pt}
\centering
\begin{subfigure}{0.5\linewidth}
    \includegraphics[width=\textwidth]{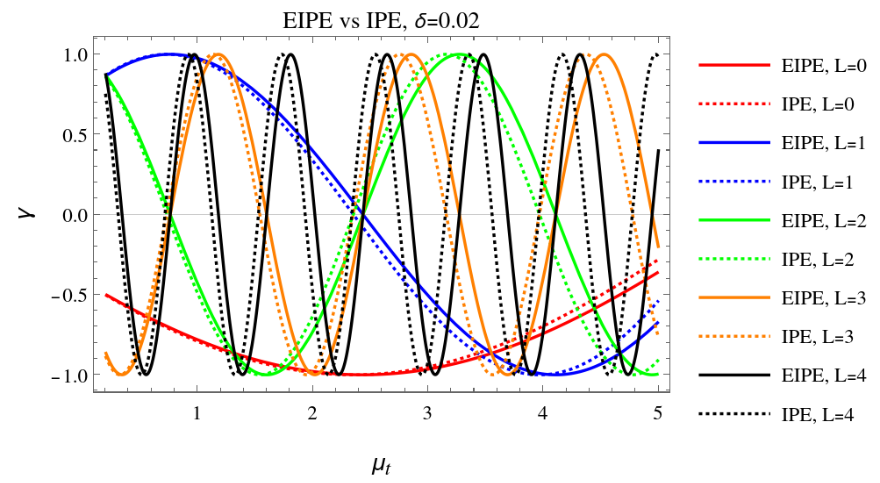}
    \caption{}
    \label{fig:eipe-vs-ipe-fixed-d}
\end{subfigure}
\begin{subfigure}{0.445\linewidth}
    \includegraphics[width=\textwidth]{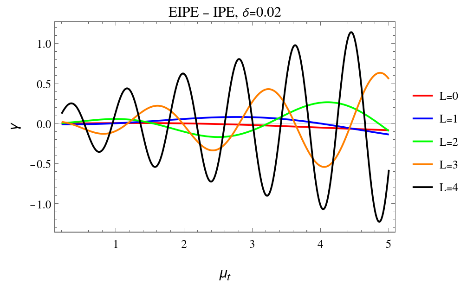}
    \caption{}
    \label{fig:error-eipe-vs-ipe-fixed-d}
\end{subfigure}
\begin{subfigure}{0.5\linewidth}
    \includegraphics[width=\textwidth]{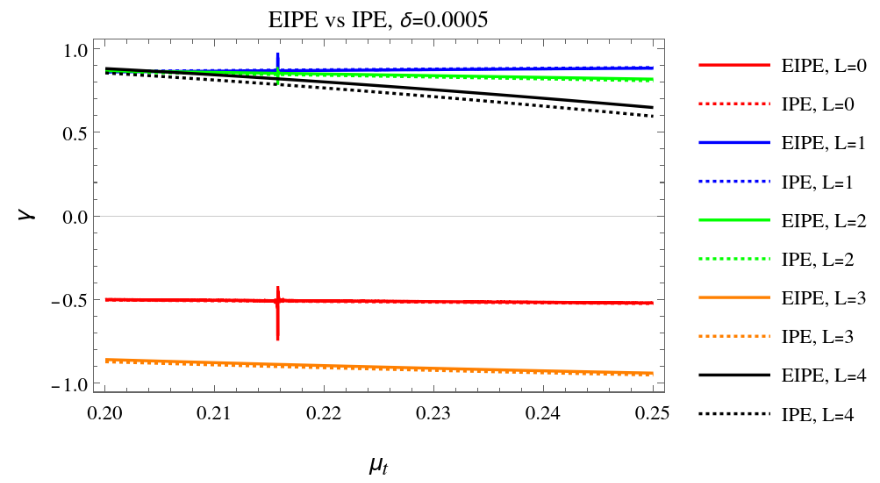}
    \caption{}
    \label{fig:eipe-vs-ipe-fixed-d-small}
\end{subfigure}
\begin{subfigure}{0.445\linewidth}
    \includegraphics[width=\textwidth]{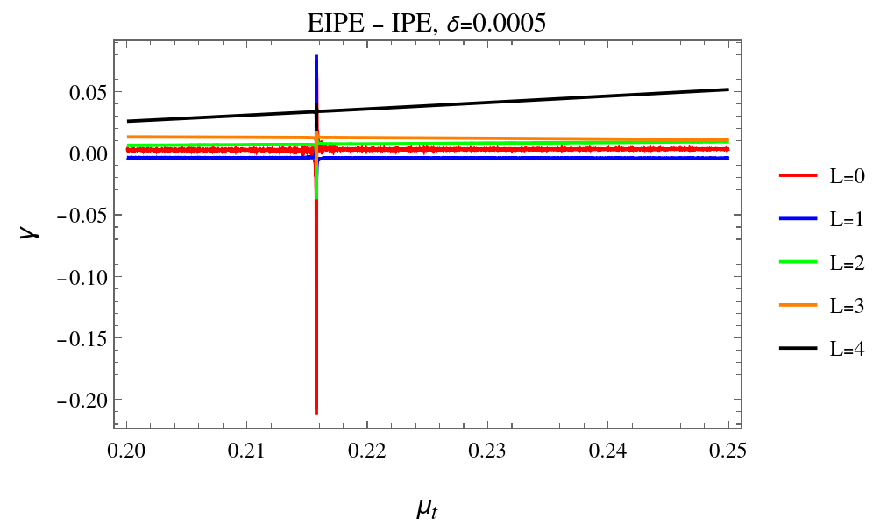}
    \caption{}
    \label{fig:error-eipe-vs-ipe-fixed-d-small}
\end{subfigure}
\begin{subfigure}{0.5\linewidth}
    \includegraphics[width=\textwidth]{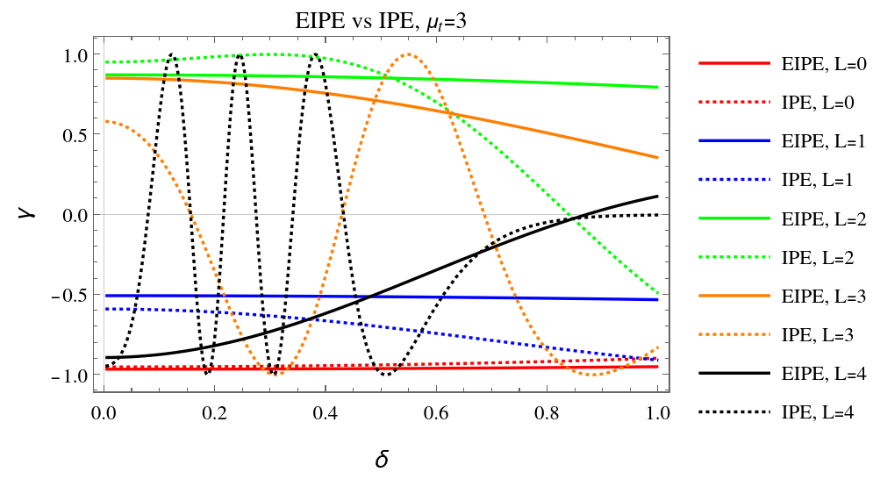}
    \caption{}
    \label{fig:eipe-vs-ipe-fixed-t}
\end{subfigure}
\begin{subfigure}{0.445\linewidth}
    \includegraphics[width=\textwidth]{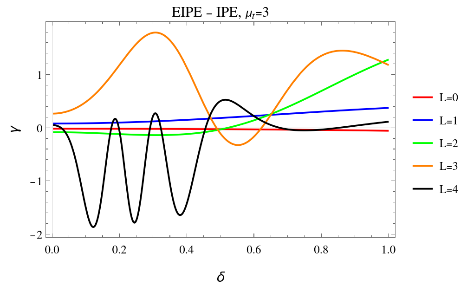}
    \caption{}
    \label{fig:error-eipe-vs-ipe-fixed-t}
\end{subfigure}
\vspace{-5pt}
\caption{Numerical comparison between the IPE and our EIPE. (a) EIPE vs IPE for different values of $\mu_t$ and (b) their difference. (c) EIPE vs IPE with respect to the length of the frustum $\delta_i$ and (d) their difference.}
\label{fig:eipe-ipe}
\end{figure*}
\section{Numerical Analysis between IPE and EIPE}
\noindent
We compare the exact value of the EIPE with the approximation in \cref{eq:approx-ipe} used by mip-NeRF \cite{barron2021mip}. In \cref{fig:eipe-vs-ipe-fixed-d} we contrast the value of the EIPE vs the IPE for frustums of length $\delta_i = 0.02$ at different positions along the ray $\mathbf{d}$ and at different positional encoding frequencies $L$. The values of $\mathbf{d}$, $\mathbf{o}$ and $\mathbf{R}$ correspond to a random pixel of a random image of the blender dataset. It is seen that the approximation is precise for frustums that are near the camera (small $\mu_t$), but it degrades the further it gets. It is also observed that this effect grows faster for larger values of $L$. This trend is more noticeable in the plot of the error between the EIPE and IPE (\cref{fig:error-eipe-vs-ipe-fixed-d}), where the magnitude of the error is a periodic function approximately bounded by two lines whose pendant seems to grow proportional with $L$. Furthermore, it is observed that the frequency of the error is also proportional to $L$. \cref{fig:eipe-vs-ipe-fixed-d-small,fig:error-eipe-vs-ipe-fixed-d-small} show a similar analysis for small values of $\mu_t$ and $\delta_i=5\times10^{-4}$, which correspond to small frustums. In these instances, it is observed that numerical errors occur, which is consistent with the analysis of the \textit{Impact of Numerical Underflow} in \cref{sec:results}. A similar analysis for a fixed value of $\mu_t=3$ and varying $\delta_i$ is shown in Figs. \ref{fig:eipe-vs-ipe-fixed-t} and \ref{fig:error-eipe-vs-ipe-fixed-t}. Here, a more drastic error is seen when $\delta_i$ increases, which is consistent with the observation made in \cite{barron2022mip} that the IPE does not approximate well for very elongated Gaussians. Additionally, rapid changes in the IPE are observed for small variations in the length of the frustum (see \cref{fig:eipe-vs-ipe-fixed-t} IPE $L=3$ and IPE $L=4$), which might not be desired. On the other hand, our EIPE is more robust to these elongations, meaning that it could be a more reliable parameterization for distant objects. 

Despite the increasing error in the approximation of the IPE for larger values of $L$, this effect gets mitigated by the nature of the IPE itself, which gives more importance to the components of the positional encoding with smaller frequencies. However, in scenarios with distant backgrounds where more elongated frustum arise, such as in the bicycle scene, {\name} seems to perform better (\cref{sec:results}). Given that the scenes in the blender and mip-NeRF 360 datasets are composed of one central object only, it is difficult to evaluate the performance of the IPE and EIPE formulations for distant objects or scenarios with several objects.

\begin{figure*}[t]
\centering
\subfloat{\includegraphics[width=\linewidth]{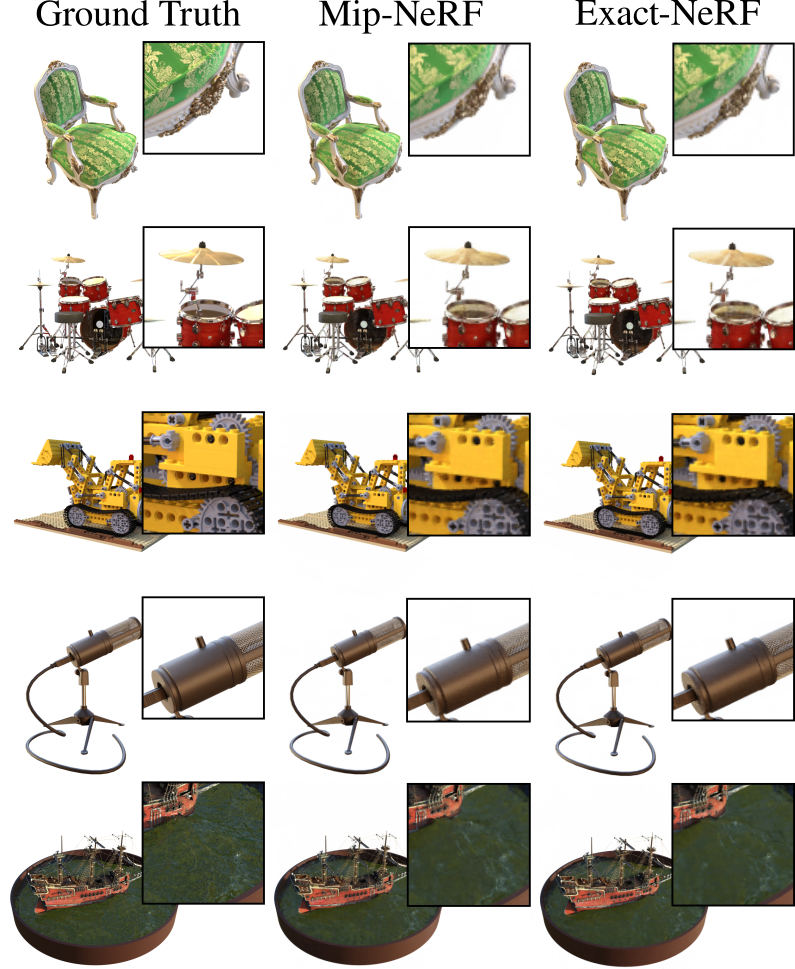}}
\caption{Additional results of {\name} for the blender dataset.}
\label{fig:lo}
\end{figure*}
\section{Additional Results on the Blender Dataset}
\noindent
We present more qualitative comparisons in \cref{fig:lo} between different scenes of the blender dataset. The reconstructions of both mip-NeRF and {\name} are almost identical, but a few differences can be noted, \eg, the apron of the chair and the holes in the lego scene are slightly sharper in our reconstruction; the details in the cymbals of the drum are more similar with the ground truth; the reconstruction of the water in the ship scene is more accurate with our method. Besides these minimal differences, our exploratory work demonstrates that analytical solutions to a volumetric positional encoding exist if the shape of the frustum is changed.

\section{Limitation of Existing Metrics}
\noindent
Following the approach from previous NeRF research, we report PSNR, SSIM and LPIPS as our evaluation metrics. PSNR and SSIM are two of the first evaluation metrics for image reconstruction. Traditionally, PSNR (based on the MSE metric) has been used to assess the quality of lossy compression algorithms. Since the PSNR is obtained via the pixel-wise absolute error, it cannot measure the structural and/or perceptual similarity between the reconstructed and reference images. SSIM was proposed as an alternative metric since it quantifies the relation between the pixels and their neighbourhood (\ie, the structural information). Several works have focused on the weakness of these metrics  \cite{wang2004, mse-crit-1-2009, kotevski2009, sharif2018suitability, ndajah2010ssim}, where the main criticism is that images subject to different compression artifacts and distortion effects (such as additive Gaussian blurring) exhibit similar PSNR and SSIM values. Additional work \cite{hore2010} has shown analytical and experimental relations between both metrics, meaning that they are not independent. In order to overcome these effects, recent image quality assessment methods have been proposed. Ding \etal \cite{ding2021comparison} have carried out a comprehensive comparison between different metrics, where deep neural networks-based metrics such as LPIPS \cite{zhang2018perceptual} and DISTS \cite{ding2020image} showed to be the most reliable quality metrics for perceptual similarity. These metrics compare two images by measuring the distance of their feature maps from a pretrained neural network. These results motivated us to include the DISTS metric in our experiments (\cref{Table:results-mipnerf,Table:results-mipnerf360}). Our method obtains a better performance in the LPIPS and DISTS metrics, thus improving the perceptual quality.

\end{document}